\documentclass[jmlr]{colt2012}  

\usepackage{times}
\usepackage{amsfonts}
\usepackage{latexsym}
\usepackage{color}
\usepackage{graphics}
\usepackage{enumerate}
\usepackage{amstext}

\usepackage{picins}
\usepackage{wrapfig}
\usepackage{amstext}
\usepackage{bbm}
\usepackage{xspace}

\title[Graph Laplacians on Singular Manifolds]{Toward understanding complex spaces: graph Laplacians on manifolds with singularities and boundaries}

\author{\Name{Mikhail Belkin} \Email{\tt mbelkin@cse.ohio-state.edu}\\
\Name{Qichao Que} \Email{\tt que@cse.ohio-state.edu}\\
\Name{Yusu Wang} \Email{\tt yusu@cse.ohio-state.edu}\\
 \addr The Ohio State University, Columbus, OH 43210, USA.\\
 \Name{Xueyuan Zhou} \Email{\tt zhouxy@uchicago.edu}\\
 \addr Department of Computer Science, University of Chicago, USA.
 }

\date{}

\def\R{\mathbb{R}}
\def\B{\mathbb{B}}

\def\vol{\mathrm{vol}}

\def\d{\partial}
\def\n{\textrm{\bf{n}}}

\def\M{\mathcal{M}}

\newcommand{\Om}{\overline{\Omega}}
\newcommand{\om}{\Omega}
\newcommand{\E}{\mathbb{E}}

\newcommand{\corner}    {edge-type\xspace} 
\newcommand{\Corner}    {Edge-type\xspace} 
\newcommand{\bndS}  {boundary-type\xspace} 
\newcommand{\BndS}  {Boundary-type\xspace} 
\newcommand{\intersection}  {intersection-type\xspace} 
\newcommand{\Intersection}  {Intersection-type\xspace} 
\newcommand{\regular}   {{regular\xspace}}

\newcommand{\myL}   {{L}}

\newcommand{\myT}[2]  {T_{#1,#2}}

\newcommand{\yusuremove}[1] {{}}

\begin{document}

\maketitle

\vskip-.5in
\begin{abstract}
In manifold learning, algorithms based on graph Laplacian constructed from data have received considerable attention both in practical applications and theoretical analysis. Much of the existing work has been done under the assumption that the data is sampled from a manifold without boundaries and singularities or that
the functions of interest are evaluated away from such points.

At the same time, it can be argued that singularities and boundaries are an important aspect of the geometry of realistic data. Boundaries occur whenever the process generating data has a bounding constraint; while singularities appear when two different manifolds intersect or if a process undergoes a ``phase transition'', changing non-smoothly as a function of a parameter.

In this paper we consider the behavior of graph Laplacians at points at or near boundaries and two main types of other singularities: \emph{intersections}, where different manifolds come together and sharp \emph{``edges''}, where a manifold sharply changes direction. We show that the behavior of graph Laplacian near these singularities is quite different from that in the interior of the manifolds. In fact, a phenomenon somewhat reminiscent of the Gibbs effect in the analysis of Fourier series, can be observed in the behavior of graph Laplacian near such points. Unlike in the interior of the domain, where graph Laplacian converges to the Laplace-Beltrami operator, near singularities graph Laplacian tends to a first-order differential operator, which exhibits different scaling behavior as a function of the kernel width. One important implication is that while points near the singularities occupy only a small part of the total volume, the difference in scaling results in a disproportionately large contribution to the total behavior. Another significant finding is that while the scaling behavior of the operator is the same near different types of singularities, they are very distinct  at a  more refined level of analysis.

We believe that a comprehensive understanding of these structures in addition to the standard case of a smooth manifold can take us a long way toward better methods for analysis of complex non-linear data and can lead to significant progress in algorithm design.
\end{abstract}




\vspace*{0.1in}
\section{Introduction}

Dealing with high-dimensional non-linear  data is one of the key challenges in modern data analysis. In recent years a class of methods based on the mathematical notion of a manifold has become popular in machine learning, starting with the papers~\cite{Roweis2000,Tenenbaum2000}. The underlying intuition is that a process with a small number of parameters will generate a low-dimensional surface in the potentially very high-dimensional space of features and that this situation is ubiquitous in real-world data. This idea is captured nicely by the technical notion of a smooth embedded Riemannian  manifold,
which provides the first realistic model for general non-linear data. Still it does not reflect certain important aspects of real data, which can be mathematically understood as singularities and boundaries.

\parpic[r]{\includegraphics[height=2.5cm,width=3cm]{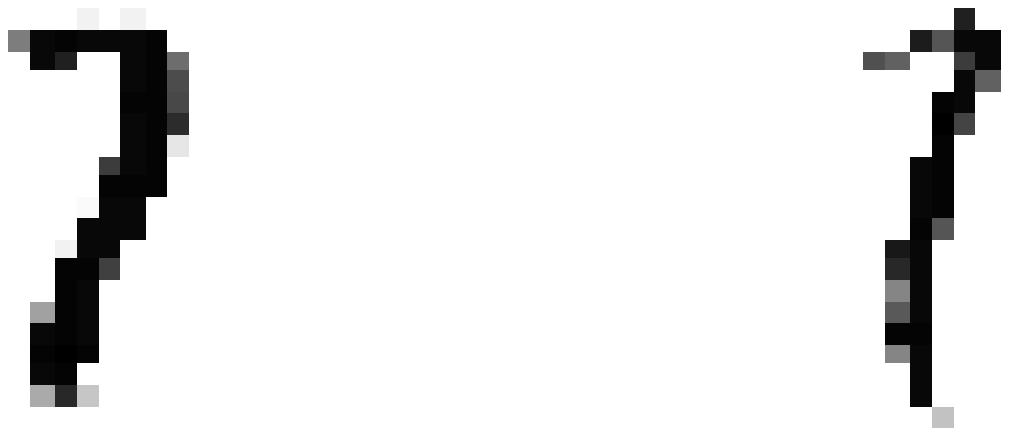}}
The most basic and, arguably, the most important singularity in real data is an {\it intersection}, where two different manifolds come together in a region of  space. This often happens in classification where two classes with presumably different structure can give rise to similar objects (consider, e.g., the similarity  of  MNIST digits ``7'' and ``1'' on the right). Another important type of singularity is a (co-dimension one)  ``{\it edge}'', which is ubiquitous in computer graphics (think of an edge of the surface of a table) and happens whenever the behavior of an underlying process changes rapidly beyond a certain point (a ``phase transition''). Finally, {\it boundaries} occur when there are bounding constraints on  the underlying process (think how poses of the human body are constrained by the range of motion of the joints) or  on the representation in the ambient space (e.g., pixels cannot have negative intensity).


In this paper we provide an analysis of these three cases for Laplacian-based learning algorithms. It turns out that all of these singularities
result in a behavior near the singularity markedly different from that inside the manifold, a phenomenon somewhat reminiscent of the Gibbs effect in Fourier analysis. In particular, the scaling of the operator is different near singularities. As a result of this scaling behavior, singularities cannot be ignored, even globally, despite the fact that relatively few points are located near them, since each of these points contributes disproportionately to the total operator. We also find that at a finer level of analysis these singularities have quite different effects, which are discussed below.

We believe that a comprehensive understanding of these structures in addition to the standard case of a smooth manifold can take us a long way toward better methods for analysis of complex non-linear data and can lead to significant progress in algorithm design.

\vspace*{0.05in}
\noindent{ \bf Related work.} Methods based on graph Laplacians constructed from data have gained acceptance for a range of inference tasks including clustering~\cite{uvon}, semi-supervised learning (e.g. ~\cite{chapelle2006ssl,zhu2006semi}) and dimensionality reduction~\cite{BelkinLapMap2003}, as well as others. An analysis in~\cite{BelkinLapMap2003} provided a mathematical framework for many of these methods by connecting graph Laplacian constructed from data to the Laplace-Beltrami operator of the underlying manifold based on the relation of the Laplacian and the heat equation. That analysis was extended and generalized in~\cite{lafon,hein,CoifmanLafon2006,belkin2008, singer,gine,Hein07graphlaplacians,CLEM_08} providing a detailed understanding of graph Laplacians obtained from manifold data. While boundary effects have been studied in non-parametric kernel smoothing (see, e.g.,~\cite[Chapter 4.4]{hardle}), they have not been considered in the Laplacian-related literature, although we note that boundary behavior for the infinite data case can be derived from the Taylor series expansion in  Lemma~9 of~\cite{CoifmanLafon2006}. To the best of our knowledge, other singularities are not considered in that literature. (In fact, the reason causing the different scaling behavior for intersection and edge types of singularities is somewhat different from that for the boundary case.)

In related developments, a considerable amount of recent work has been aimed at understanding the case of intersecting linear spaces using various techniques from
algebraic geometry to spectral clustering (e.g.,~\cite{vidal2005generalized,chen2009foundations,LerZha2010}) both in terms of algorithms and theoretical guarantees. In our setting, this is a special case of a singular manifold with intersection singularities. Another related work is~\cite{multimanifold}, where the authors analyze a model based on a mixture of manifolds in the context of semi-supervised learning.

We also note that there has been much recent interest in reconstructing topological invariants of manifolds and other spaces (see, e.g, \cite{CCL09,CO08,NSW08,NSW11}).
The line of work most related to our results is on learning stratified spaces, where multiple submanifolds (strata) are ``glued'' nicely together~\cite{ACC11,BenWangMuk1,HRS08, Bendich:2007}, which provides an even more general model for a singular space.

\vspace*{0.05in}
\noindent{\bf Summary of results.} Consider the (appropriately scaled, see Section~2) graph Laplacian $\myL_{n,t}$ constructed from $n$ data points, using the Gaussian kernel with bandwidth $t$. It can be shown (see the references above) that as $t$ tends to zero and $n$ tends to infinity at an appropriate rate, $\myL_{n,t}$ converges to the Laplace-Beltrami operator $\Delta$  (a second order differential operator) inside the manifold. In this paper, we show that for singular manifolds $\myL_{n,t}$ exhibits a very different behavior near the singularity set. Specifically,  for a sufficiently smooth function $f$ and a small $t$, within distance $\sqrt{t}$ of the singularity set, $\myL_{n,t} f$ is approximated by $\frac{1}{\sqrt t} D$, where $D$ is
some first order derivative operator that will be described explicitly. The difference in scaling becomes crucial when $t$ is small as the $\frac{1}{\sqrt t}D$ term becomes dominant. To simplify the discussion, we will first assume that the data is infinite, and  thus $\myL_{n,t} f$ can be replaced by $\myL_{t} f$. Finite sample bounds and rates will be discussed later. The types of singularities considered in the paper are as follows:

\parpic[r]{\includegraphics[height=3cm]{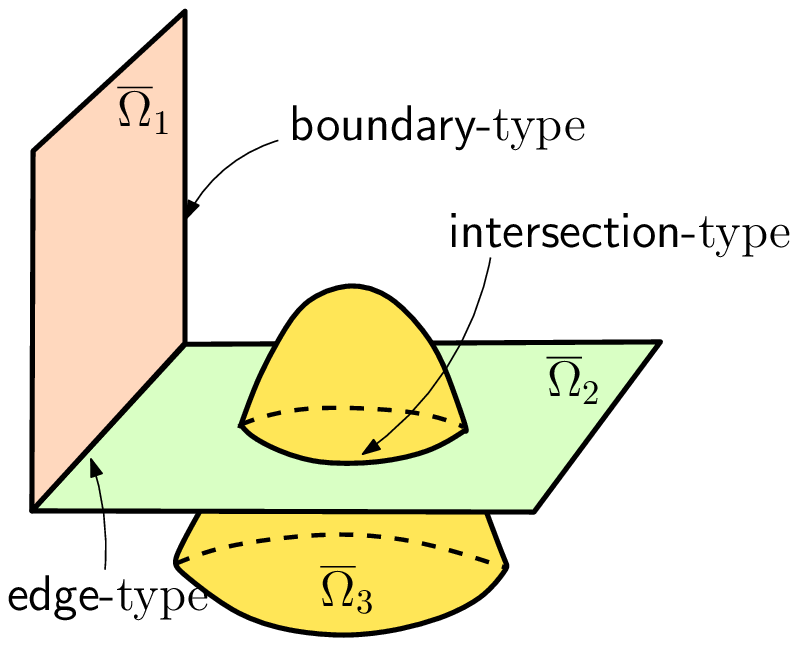}}
\noindent{\bf Boundary.} At a point $x$ near manifold boundary, $\myL_{t}$ can be approximated by $\frac{1}{\sqrt{t}}\phi(\frac{R}{\sqrt{t}})\partial_{\bf n}$, where $R$ is the distance from $x$ to the boundary and $\bf n$ is the unit vector toward the nearest boundary point (outward normal at the boundary).  $\partial_{\bf n}$ denotes the directional derivative, while $\phi(z)$ is a scalar function of the form $\phi(z) = Ce^{-z^2}$.\\
\noindent{\bf Intersection and edge.} The situation is more complicated for other types of singularities, where two different manifolds $\Omega_1$ and $\Omega_2$  intersect or come together. Given a point $x_1 \in \Omega_1$ consider its projection $x_2$  onto $\Omega_2$ and its nearest neighbor $x_0$ in the singularity.
Similarly to the boundary case, let ${\bf n}_1$ and ${\bf n}_2$, be the directions to $x_0$ from $x_1$ and $x_2$ respectively and let $R_1$ and $R_2$ be the corresponding distances.
$\myL_{t}f(x_1)$ can be approximated by
$\frac{1}{\sqrt{t}}\phi_1(\frac{R_1}{\sqrt{t}})\partial_{{\bf
n}_1}f(x_0)  +
\frac{1}{\sqrt{t}}\phi_2(\frac{R_2}{\sqrt{t}})\partial_{{\bf
n}_2}f(x_0)$.
Importantly, the form of the scalar functions $\phi_1, \phi_2$ is different for different types of singularities: for \intersection{} singularity, we have: $\phi_i (z) = A_i z e^{-Cz^2}$, while for \corner{} we have\footnote{For compactness, we are  simplifying $\phi$'s for edge singularities by removing some higher order terms (see Section~3.3).}: $\phi_i (z) = A_i z e^{-Cz^2} + B_i e^{-Cz^2}$.
Here the coefficients $A_i$, $B_i$ and $C$ depend on the angle between the manifolds and can be written explicitly.

\vspace*{0.05in}
\noindent{\bf Significant points and observations:}

\noindent{\bf 1. Scaling.}  Within $O(\sqrt{t})$ distance of the singularity, $\myL_{t}$ is dominated by a differential operator different from the Laplace-Beltrami operator.
Away from the singularity, the higher-valued (${1}/{\sqrt t}$) term fades and $\myL_{t}$  becomes an approximation to the Laplace-Beltrami operator $\Delta$.

\noindent{\bf 2. Contributions of singular points.} Let the dimension of the manifold be $d$. The volume of points within $\sqrt{t}$ of the singularity is approximately $\sqrt{t} ~\vol_{d-1} (S)$, where $\vol_{d-1} (S)$ is the $d-1$ dimensional volume of the singular points\footnote{The set of singular points $S$ is always $d-1$ dimensional (co-dimension one), for the  boundary and ``edge'' singularities (although not necessarily for the intersection).}.
Fix a function $f$. As $t \to 0$,  the integral $\myL_{t} f$ over the points within $\sqrt{t}$ of the singularity does not disappear and, in fact, will converge to a constant, depending on the function and the $d-1$ dimensional volume of the singularity $\vol_{d-1} (S)$. Moreover, because of the  scaling, the contribution of those near-singular points to the $L_2$ norm will tend to infinity.
%
We remark that, while  different from a technical point of view, this phenomenon is reminiscent to the Gibbs effect in Fourier series, where the effects of discontinuity  do not vanish as approximation becomes more  precise.

\parpic[r]{\begin{tabular}{ccc}
\includegraphics[height=2cm]{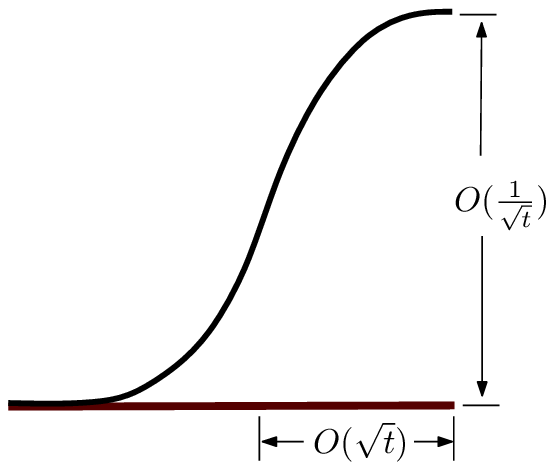} &
\includegraphics[height=2cm]{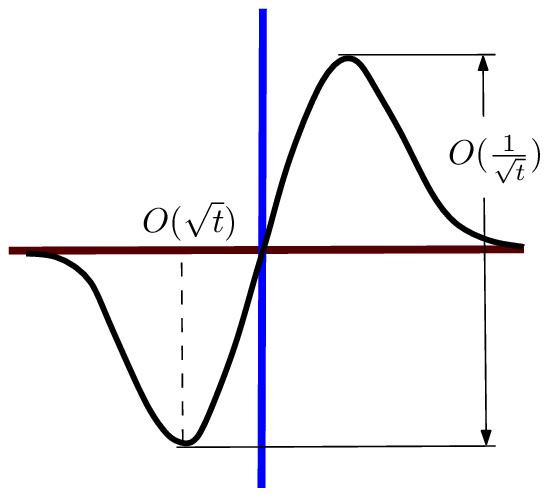} &
\includegraphics[height=2cm]{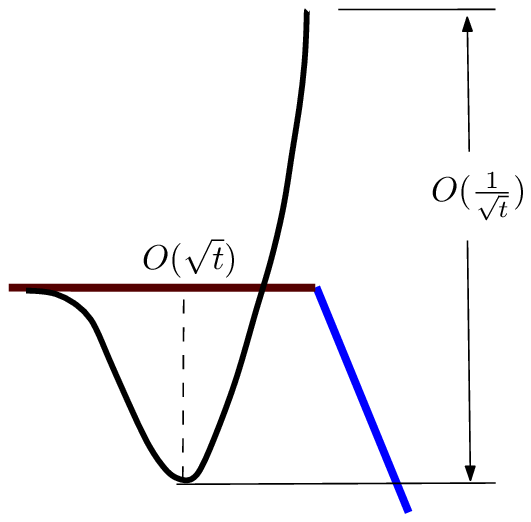} \\
(a) boundary & (b) intersection & (c) edge
\end{tabular}
}  \noindent{\bf 3. Shape near the boundary.} While the $\frac{1}{\sqrt t}$ scaling is common for all singularities, the type of singularity is reflected in the shape of the function $\phi$, which  is very different for different types of singularities. Somewhat over-simplifying matters, one can think of $\phi = e^{-z^2}$ for the boundary case, $\phi = z e^{-z^2}$ for the intersection singularity and $\phi = e^{-z^2} + ze^{-z^2}$ for the edge (see figures above). These differences  lead to quite distinctive patterns, when the operator is applied to a fixed function.
For example, the directional derivative terms disappear \emph{at} the intersection-type singular points. In fact for these points $L_t$ converges to the usual Laplace-Beltrami operator. However, at points within $\sqrt{t}$ distance of these points we expect to see $L_t f$ take both positive and negative values of magnitude $\sqrt{t}$.
We think that this difference may be a key to future algorithms.

\noindent{\bf 4. Rates and finite sample bounds.} Convergence rates for $t$ as well as finite sample bounds are provided in Section~\ref{sec:convergencerate}. Interestingly, and perhaps counter-intuitively, the scaling implies that fewer data points are required to ensure that $\myL_{t,n}$ is an accurate (relative) estimate of $L_{t}$ near the boundary than inside the domain.

\noindent{\bf 5. Impact on eigenfunctions}. Eigenfunctions of the graph Laplacian associated to data are used in a large number of applications.
While the question of convergence is very subtle (see \cite{CLEM_08} for the proof in the case of a smooth manifold), our results for fixed functions provide an indication of the expected answer assuming the convergence holds. We provide a brief discussion of it in Section \ref{sec:implication}.
It turns out that the boundary case automatically leads to Neumann boundary conditions (even though the operator does not "see" the boundary explicitly).
Quite surprisingly, the edge singularity appears to have no effect on the limit behavior of eigenfunctions,  at least if the singular manifold is isometric to a smooth one (think of bending a sheet of paper).
The case of the intersection is more complex, but intersections of co-dimension two or more seem to have no effect on the limit behavior of eigenfunctions (in $L_2$ norm).

\noindent{\bf 6. Normalized Laplacians. Multiple manifolds.} While we do not provide a detailed discussion due to space constraints, our analysis can be fairly straightforwardly generalized to other versions of graph Laplacians, including the random walk normalized, symmetric normalized, and twice normalized graph Laplacian discussed in \cite{CoifmanLafon2006,Hein07graphlaplacians}. Interestingly,  the popular symmetric normalized Laplacian does not lead to Neumann boundary conditions as a density term appears inside the directional derivative, i.e., $\sqrt{p(x)}\d_{\n} [f(x)/\sqrt{p(x)}]$. The resulting expression involves derivatives of the density and does not appear to be a natural boundary condition.

The analysis can also be extended to the case of multiple manifolds. We restrict our discussion to the case  of two manifolds to simplify the exposition.

Finally, some  experimental results can be found in Appendix \ref{appendix:sec:exp}.

\section{Problem Setup}\label{sec:setup}
Let $\Om$ be the set of $k$ smooth compact Riemannian submanifolds of intrinsic dimension\footnote{ For convenience we assume that the submanifolds have the same dimension. However, the same analysis can be applied to a  mixture of manifolds with different dimensionality after normalizing the probability density.}
$d$ embedded in $\R^N$. For each smooth component $\Om_i$, let $\om_i$ denote its interior and $\d \om_i$ the boundary of $\om_i$. We assume the boundaries $\d \om_i$ satisfy the {\em necessary smoothness conditions}\footnote{We require that  the derivatives of the first two orders of the transition map exist, i.e., $\M$ is a $C^2$-manifold}. 
Let $\om$ be the union of $\om_i$, and $\d\om$ be the union of $\d\om_i$.
$\Om_i$ and $\Om_j$ may not be disjoint, and we will consider the following two ways that they can associate with each other: they intersect in their interior (i.e, $\om_i \cap \om_j \neq \emptyset$) or they are ``glued'' along (part of) their boundaries. More precisely, given a point $x\in \Om$, we say that $x$ is \emph{\regular{}} if it falls in the interior of \emph{exactly one} submanifold $\om_i$; that is, $x\in \om_i, x\notin \om_j, j\ne i$. Otherwise, $x$ is a \emph{singular point}. There are several possible singularities, and this paper considers the following {\em three general classes}:\\
{\bf 1. \bndS{}}: Boundary points that belong to exactly one smooth submanifold $\om_i$: that is, $x\in \d\om_i, x\notin \Om_j, j \ne i$;\\
{\bf 2. \intersection{}}: Points at the intersection of two smooth submanifolds. For simplicity of exposition, we assume that the intersection happens
between the interior of the submanifolds; that is, $x \in \om_i\cap \om_j, i\ne j$;\\
{\bf 3. \corner{}}: Boundary points that belong to the boundaries of two
smooth submanifolds. That is, $x\in \d\om_i\cap \d\om_j, i\ne j$.

\vskip -0.01in For technical reasons, we will assume that the set of singular points is also a smooth manifold of lower dimension, at least locally. Moreover, for edge and intersection singularities the tangent space at each point of the singular manifold is the intersection of the tangent spaces to each ``piece''.
We will call a union of smooth manifolds $\Om$ with the above types of singularities a \emph{singular manifold}. Notice that for a \regular{} point $x$, the manifold is smooth at $x$, and the tangent space at $x$ is homeomorphic to $\R^d$.

For a singular manifold $\Om$, we consider the following \emph{piecewise-smooth function $f: \Om \mapsto \R$}. In particular, set $f_i := f |_{\Om_i}$, $i=1,\ldots, k$, be $f$ restricted to the submanifold $\Om_i$. We require that each $f_i$ is $C^2$-continuous on interior points. let $p(x)$ be a piecewise smooth probability density function on $\Om$ and we assume that $0<a\le p_i(x)\le b<\infty$.


\vspace*{0.05in}
\noindent{\bf Graph Laplacian.} Given $n$ random samples $X=\{X_1,\cdots, X_n\}$ drawn i.i.d. from a distribution with density $p(x)$ on $\Om$, we can build a weighted graph $G(V,E)$ by mapping each sample point $X_i$ to vertex $v_i$ and assigning a weight $w_{ij}$ to edge $e_{ij}$. One typical weight function is the Gaussian, which is used in this paper and defined as follows:
\vspace*{-0.10in}\begin{equation*}
    w_{n,t}(X_i,X_j)=\frac{1}{nt}K_t(X_i,X_j)=\frac{1}{n}\frac{1}{t^{d/2+1}}e^{-\frac{\|X_i-X_j\|^2_{\R^N}}{t}}.
\end{equation*}
Notice that in this Gaussian weight function, the Euclidean distance
is used. The normalization by $\frac{1}{nt^{d/2+1}}$ is for the
convenience of limit analysis.

Let $W_{n,t}$ be the edge weight matrix of graph $G$ with $W_{n,t}(i,j)=w_{n,t}(X_i,X_j)$, and $D_{n,t}$ be a diagonal matrix such that $D_{n,t}(i,i)=\sum_{j}w_{n,t}(X_i,X_j)$, then the unnormalized graph Laplacian is defined as the $n\times n$ matrix $\myL_{n,t} = D_{n,t} - W_{n,t}.$ Now for a fixed smooth function $f(x)$, and any point $x\in \Om$,
the graph Laplacian applied to this function $f$ is thus
\vspace*{-0.1in}\begin{equation*}
    \myL_{n,t} f(x)=\frac{1}{nt}\sum_{j=1}^n K_t(x,X_j)[ f(x)-f(X_j) ].
\end{equation*}
{\bf Limit behavior of graph Laplacian.} The limit study of graph Laplacians primarily involves the limits of two parameters, sample size $n$ and weight function bandwidth $t$. As $n$ increases, one typically decreases $t$ to let the graph Laplacian capture progressively a finer local structure. With a proper rate as a function of $n$ and $t$, the limit of $\myL_{n,t}$ and its various aspects at \regular{} points, including the finite sample analysis, are studied in \cite{belkinThesis,lafon,hein,singer,gine,Hein07graphlaplacians,belkin2008,CLEM_08}. The basic result is that the limit of $\myL_{n,t} f(x)$ at a \regular{} point $x$ is $\myL_{n,t}f(x)\stackrel{p}{\to} -\frac{1}{2}\pi^{d/2}p(x)\Delta_{p^2} f(x)=-\frac{1}{2}\pi^{d/2}p(x)\{\frac{1}{p^2}\textrm{div}[p^2\textrm{grad f(x)}]\}$, where $\Delta_{p^2}$ is called the {\em weighted Laplacian}, see \cite{grigoryan2006}. See \cite{Hein07graphlaplacians} for the limit analysis of other versions of graph Laplacians at regular points.

\section{Limit Analysis of Graph Laplacian on Singular
Manifolds}\label{sec:GraphLaplacianLimit}



The limit analysis of graph Laplacians typically involves limits of two parameters, the sample size $n$ and  the (Gaussian) kernel bandwidth $t$. This is usually done in two steps, first analyzing the limit $t\to 0$ for infinite data and then obtaining finite sample results and rates as $n \to \infty$ using concentration inequalities.

In this section we analyze the behavior of the infinite graph Laplacian (i.e., $\myL_{n,t}, n=\infty$),  $\myL_{t} f(x)$, when $x$ is on or near a singular point, $t$ is small and the function  $f:\Om\rightarrow \R$, is fixed. Finite sample results and rates are given in Section \ref{sec:convergencerate}.



For a fixed $t$ we define $\myL_{t}$ as the limit of $\myL_{n,t}$ as the amount of data tends to infinity:
%
\vspace*{-0.10in}\begin{equation}\label{eqn:aimintegral}
\myL_{t} f(x) =  \myL_{\infty,t}f(x) = {\E_{p(X)} [\myL_{n,t} f(x)]}
f(x)=\frac{1}{t}\int_{\Om}K_t(x,y)(f(x)-f(y))p(y)dy.
\end{equation}

\noindent {\bf \em Local coordinate system}: The above integral is defined on the manifold $\Om$. In order to study the behavior of this integral for a small $t$, we need to introduce a  local coordinate systems at $x$. The most convenient coordinate system for our purposes\footnote{We note that another possibility is to use the coordinate system given by the exponential map (as in~\cite{belkinThesis,CoifmanLafon2006}). This has an advantage of being independent of the embedding, but requires more subtle analysis in the presence of singularities.} is obtained by a local projection from the manifold to its tangent space $\pi_x:\Om \to T_x$.
This projection is one-to-one and smooth for points sufficiently close to $x$.
Hence its inverse, $\pi_x^{-1}$ exists as well.
Note that  on an intersection or edge singularity the projection is onto the union of tangent spaces. However, it is still one-to-one as long as we restrict the projection of points from each piece to their respective tangent space.

In what follows, we often use change of variables to convert an integral over a manifold to an integral over the tangent space at a specific point. The following bounds are used throughout the paper. For two points $x, y \in \Om_i$, let $y' = \pi_x (y)$ be the projection of $y$ in the tangent space $T_x$ of $\Om_i$ at $x$. Let $J \pi_x |_y$ (resp. $J \pi^{-1} |_{y'}$) denote the Jacobian of the map $\pi_x$ at point $y \in \Om_i$ (resp. of the inverse map $\pi^{-1}_x$ at $y' \in T_x$). For $y$ sufficiently close to $x$, we have (e.g., \cite{NSW08})
\vspace*{-0.10in}\begin{align}
\| x - y \| = \| x - y' \| + O( \| x - y' \|^3) &\Rightarrow \|x - y \|^2 = \| x - y' \|^2 + O(\| x - y'\|^4), \label{equ:basicApprox1} \\
|J \pi_x |_y - 1 |= O(\|x - y\|^2)~~&\text{and}~~| J\pi^{-1}|_{y'} - 1 | = O(\|x - y' \|^2).
\label{eqn:jacobianapprox}
\end{align}

\subsection{\BndS{}: Manifold with Boundary}
We first analyze the limit of $\myL_{t} f(x)$ when $x$ is near a {\em smooth}
boundary of a single smooth manifold $\Om$ as the results for the boundary case are simpler to state.
\begin{theorem}
\label{thm:bndS} Let $f\in C^2(\Om)$, $p(x)\in C^\infty(\Om)$ with $0<a\le p(x)\le b<\infty$, and $\d \Omega$ be the smooth boundary of $\Omega$. Given a point $x$ near the boundary, let $x_0$ be its nearest neighbor on the boundary $\d\om$, and $\n$ the inward normal direction of $\Omega$ at $x_0$.
Put $\|x-x_0\|=r\sqrt{t}$. For $t$ sufficiently small  we have
\vspace*{-0.10in}\begin{equation}
\myL_{t} f(x) = -  \frac{1}{\sqrt{t}} \frac{\pi^{(d-1)/2}}{2}e^{-r^2}  p(x_0)\d_\n
f(x_0)+o\left(\frac{1}{\sqrt{t}}\right) \label{eqn:bndNeighbor}.
\end{equation}

Consequently, if $x$ is on the boundary, i.e., $r=0$, then $x = x_0$ and
\vspace*{-0.10in}\[
\myL_{t} f(x) = -  \frac{1}{\sqrt{t}} \frac{\pi^{(d-1)/2}}{2} p(x)\d_\n
f(x)+o\left(\frac{1}{\sqrt{t}}\right).
\]
\end{theorem}

The proof of the theorem can be found in Appendix \ref{appendix:sec:bnd}.
For comparison, the corresponding result for interior points is as follows:
\vspace*{-0.15in}\begin{equation}
\myL_{t} f(x) = - \frac{1}{2}\pi^{d/2} p(x) \Delta_{p^2} f(x)+o(1)\label{eqn:interior},
\end{equation}
Hence the graph Laplacian has
a different behavior on or near a boundary point from interior
points. The values of $\myL_t f(x)$ are of different
order, $O(1/{\sqrt{t}})$ near the boundary as opposed to $O(1)$ in
the interior.
Intuitively, the scaling difference stems from integrating over a half-plane for boundary points versus integrating over a (high-dimensional) plane for points away from boundary, which causes the integration of the first-order derivative terms non-vanishing in the former case.

In practice, we do not know the boundary, and apply
the same {\em global} normalization for all $x\in \Om$. The result
is that the large values of $\myL_tf(x)$ are likely to correspond to
points near the boundary (or other singular set, see below).

We think that these observation could lead to useful techniques for data analysis and  algorithms that respect singularities.
See Appendix \ref{appendix:sec:expdetection} for some  experiments on  MNIST dataset.

\subsection{\Intersection: Intersection of Manifolds}\label{sec:inter}
We now present results on the behavior of the graph Laplacian near the intersection of two $d$-manifolds $\Om_1$ and $\Om_2$ embedded in $\R^N$ with $N > d$. Note that we do not assume that the intersection is of codimension one.
We also remark that while the boundary behavior is caused by the integration over a half-disk (which causes asymmetry in the integration of first-order terms) for boundary points, this is not the case for points around intersection singularity and edge-type singularity.
The behavior around these latter two types of singularities are more involved and are more subtle to analyze.

\begin{theorem}\label{thm:intersection}
Let $\Om_1$ and $\Om_2$ be two $d$-dimensional smooth manifolds in $\R^N$ potentially with boundaries, and their intersection $\om_1 \cap \om_2$ is a smooth manifold of dimension $l(\leq d-1)$. Let $f$ be a continuous function over $\Om = \Om_1 \cup \Om_2$ whose restriction $f_i: = f | _{\Om_i}$ on $\Om_i$, $i = 1,2$, is $C^2$-continuous. Given a point $x\in \om_1$ near the intersection, let $x_0$ be its nearest neighbor in $\om_1 \cap \om_2$, and $x_1$ (resp. $x_2$) be its projection in the tangent space of $\Om_1$ at $x_0$ (resp. in the tangent space of $\Om_2$ at $x_0$). Put $\| x - x_0 \| = r \sqrt{t}$. For a sufficiently small $t$, we have
\vspace*{-0.10in}\begin{equation}\label{eqn:intNeighbor}
\myL_{t} f(x) = \frac{1}{\sqrt{t}} \pi^{d/2} re^{-r^2\sin^2\theta }
p(x_0)(\d_{\n_1} f_1(x_0)+\cos\theta \d_{\n_2}
f_2(x_0))+o\left(\frac{1}{\sqrt{t}}\right),
\end{equation}
where  $\n_1$ and $\n_2$ are the unit vectors in the direction of $x_0 - x_1$ and $x_0 - x_2$, respectively, and $\theta$ is the angle between $\n_1$ and $\n_2$.
\end{theorem}
\parpic[r]{\includegraphics[height=3cm]{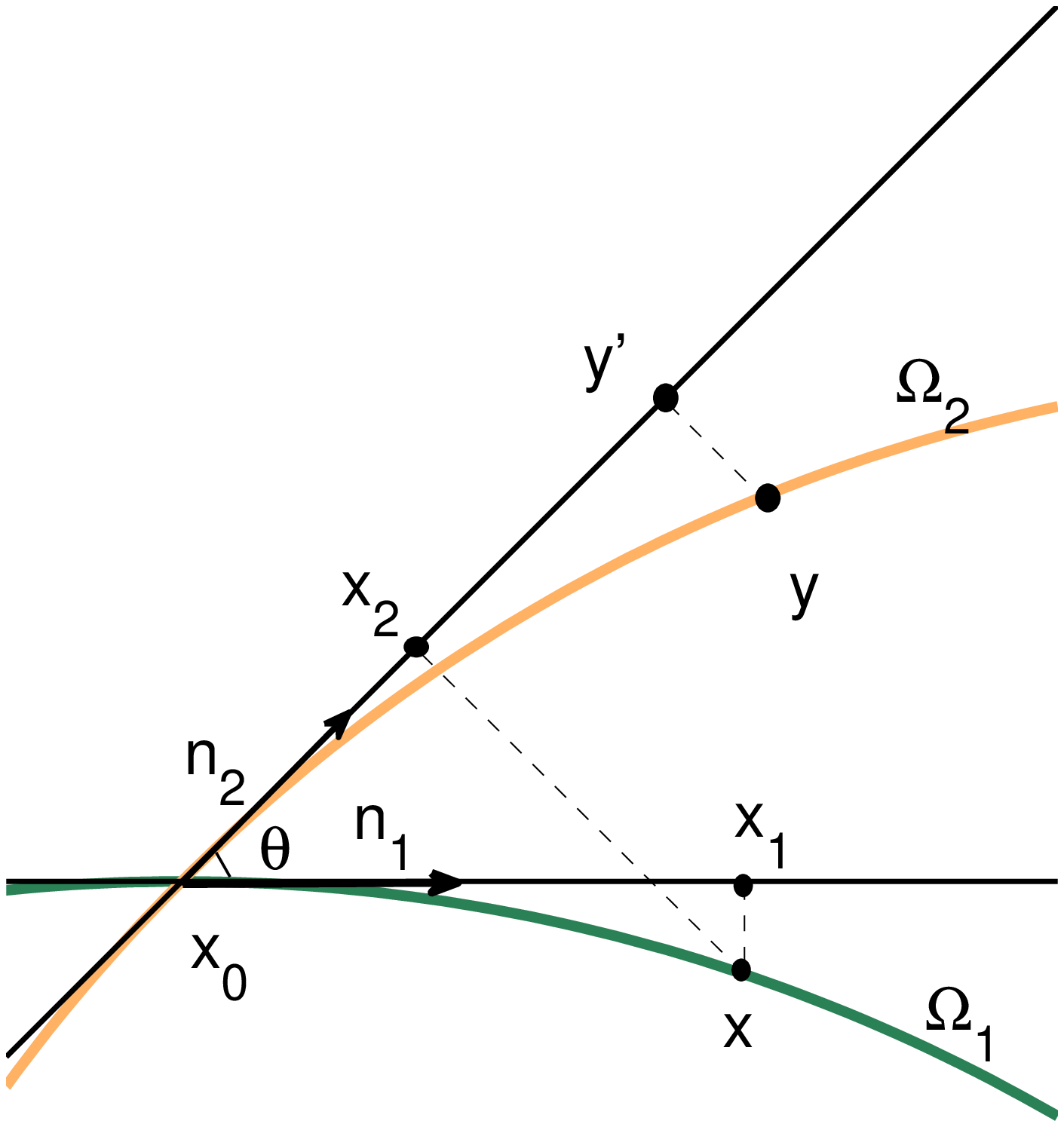}}
\noindent {\bf Proof.~}
Given $x\in \om_1$ and its nearest neighbor $x_0$ in $\om_1\cap \om_2$, recall that $x_1$
and $x_2$ are the
orthogonal projection of $x$ onto the tangent space $\myT{x_0}{\om_1}$ of $\om_1$ at $x_0$, and onto the tangent space $\myT{x_0}{\om_2}$ of $\om_2$ at $x_0$, respectively.
For any $y \in \om_2$, let $y'$ denote its orthogonal projection onto the tangent space $\myT{x_0}{\om_2}$. See the right figure for an illustration.
By the definition of $\myL_{t}$, we have $\myL_{t} f(x) = \frac{1}{t}\int_{\Om} K_t(x,y)(f(x)-f(y))p(y)dy$. Since $\Om = \Om_1\cup \Om2$, this can then be decomposed as

\begin{align}
\myL_{t} f(x) 
& = \frac{1}{t}\int_{\Om_1} K_t(x,y)(f_1(x)-f_1(y))p(y)dy+
\frac{1}{t}\int_{\Om_2} K_t(x,y)(f_1(x)-f_2(y))p(y)dy \label{eqn:Lt:int}
\end{align}

The first term above, which is the integral over $\Om_1$, is exactly the graph Laplacian of $f_1$ at an interior point $x$ of $\Om_1$. Thus, it is bounded by $O(1)$. We now focus on the second integral over $\Om_2$.
Let $\B(x)$ denote the ball in $\R^N$ center around $x$ with radius $t^{\frac{1}{2}-\varepsilon}$ for a sufficiently small constant $\varepsilon>0$.
In the derivation below, we will approximate the integral over the manifold $\Om_2$ by an integral over the region $\B(x) \cap \Om_2 \subseteq \Om_2$. The error term induced is $O(e^{-t^{-\varepsilon}}) = o(1)$ for small $t$.
Using the fact that $f_1(x_0) = f_2(x_0)$, we have:
\begin{align}
 & \frac{1}{t}\int_{\Om_2} K_t(x,y)(f_1(x)-f_2(y))p(y)dy ~=~ \frac{1}{t}\int_{\B(x)\cap \Om_2}K_t(x,y)(f_1(x)-f_2(y))p(y)dy + o(1) \nonumber \\
=& \frac{1}{t}\int_{\B(x)\cap \Om_2}K_t(x,y)(f_1(x)-f_1(x_0))p(y)dy+\frac{1}{t}\int_{\B(x)\cap\Om_2}K_t(x,y)(f_2(x_0)-f_2(y))p(y)dy + o(1) \nonumber \\
=&
\frac{1}{t}(f(x)-f(x_0))\int_{\B(x)\cap \Om_2}K_t(x,y)p(y)dy+\frac{1}{t}\int_{\B(x)\cap\Om_2}K_t(x,y)(f(x_0)-f(y))p(y)dy + o(1).
\label{eqn:A1}
\end{align}

Since $\| x - x_0 \| = r\sqrt{t}$, it follows from Eqn (\ref{equ:basicApprox1}) that $\| x_0 - x_1 \| = r\sqrt{t} + O(t^{3/2})$.
It can then be shown that $\|x-x_2\| = r \sqrt{t} \cdot \sin{\theta}+{O}(t)$ and $\|x_0 - x_2 \| = r \sqrt{t} \cdot |\cos{\theta}| + O(t)$.
For simplicity, let $\tilde{O}(t^\beta)$ denote $O(t^{\beta - \epsilon})$ for a sufficiently small positive constant $\epsilon$.
For a point $y \in \B(x) \cap \Om_2$ and its projection $y'$ in $\myT{x_0}{\om_2}$,
we have that $\| x - y \| = \tilde{O}(t^{\frac{1}{2}})$ and $\|x_0 - y \| \le \|x_0 - x \| + \|x - y \| = \tilde{O}(t^{\frac{1}{2}})$.
This implies that $\|y - y'\| = \tilde{O}(t)$ and $\|x - y'\| = \tilde{O}(t^{\frac{1}{2}})$ by Eqn (\ref{equ:basicApprox1}). Using these distance bounds, we have:
\begin{align}
&K_t(x,y) = \frac{1}{t^{d/2}}e^{-\|x-y\|^2/t} = \frac{1}{t^{d/2}}e^{-\|x-y' + y' - y\|^2/t}
= \frac{1}{t^{d/2}}e^{-\|x-y'\|^2/t} \cdot e^{-\|y' - y\|^2/t} \cdot e^{\frac{O(\|x-y'\|\cdot \|y' - y\|)}{t}} \nonumber \\
&=  \frac{1}{t^{d/2}}e^{-\|x-y'\|^2/t} \cdot (1 + \tilde{O}(t)) \cdot (1 + \tilde{O}(t^{1/2})) = \frac{1}{t^{d/2}}e^{-\|x-y'\|^2/t} \cdot (1 + \tilde{O}(t^{1/2})) \nonumber \\
&=\frac{1}{t^{d/2}}e^{-\frac{\|x-x_2\|^2}{t}+\frac{\|x_2-y'\|^2}{t}} \cdot  (1 + \tilde{O}(t^{1/2}))
=\frac{1}{t^{d/2}}(e^{-r^2\sin^2\theta}\cdot e^{O(t^{\frac{1}{2}})})e^{-\frac{\|x_2-y'\|^2}{t}} \cdot  (1 + \tilde{O}(t^{1/2})) \nonumber \\
&=\frac{1}{t^{d/2}}e^{-r^2\sin^2{\theta}}e^{-\|x_2-y'\|^2/t}(1+\tilde{O}(t^{1/2})) ~=~ e^{-r^2 \sin^2 \theta} K_t(x_2, y') (1+\tilde{O}(t^{1/2})) .\label{eqn:kernelapprox}
\end{align}

Now consider the first term in Eqn (\ref{eqn:A1}).
Let $\pi_{x_0}$ denote the projection map from $\om_2$ onto $\myT{x_0}{\om_2}$ and $\pi^{-1}$ its inverse.
Denote by $B:= \pi_{x_0}(\B(x) \cap \myT{x_0}{\om_2})$ to be the projection of $\B(x) \cap \myT{x_0}{\om_2}$ on the tangent space $\myT{x_0}{\om_2}$.
By applying Taylor expansion to $f(x)$ at $x_0$ and $p(y)$ at $x_0$, and combining Eqn \ref{equ:basicApprox1} and \ref{eqn:jacobianapprox}, we have that:
\begin{align}
&\frac{1}{t}(f(x)-f(x_0))\int_{\B(x)\cap \Om_2}K_t(x,y)p(y)dy ~=~\frac{1}{t}(\langle \nabla f(x_0), x_1 - x_0\rangle + \tilde{O}(t)) \int_{\B(x)\cap \Om_2}K_t(x,y)p(y)dy \nonumber\\
=&\frac{1}{t^{\frac{d+1}{2}}}(r\cdot \langle\nabla f(x_0),\n_1\rangle+\tilde{O}(t^{\frac{1}{2}}))\int_{B}\left[(e^{-r^2\sin^2{\theta}}e^{-\frac{\|x_2-y'\|^2}{t}}(1+\tilde{O}(t^{\frac{1}{2}})))(p(x_0)+\tilde{O}(t^{\frac{1}{2}}))\right] J \pi^{-1}|_{y'} dy' \nonumber\\
=& \frac{1}{t^{\frac{d+1}{2}}}(r \partial_{\n_1} f_1 (x_0)+\tilde{O}((t^{\frac{1}{2}}))e^{-r^2\sin^2{\theta}}(1+\tilde{O}((t^{\frac{1}{2}}))(p(x_0)+\tilde{O}((t^{\frac{1}{2}}))\int_{B}e^{-\|x_2-y'\|^2/t}dy' \nonumber \\
=& \frac{1}{t^{\frac{d+1}{2}}}(r\partial_{\n_1} f_1 (x_0)+\tilde{O}(t^{\frac{1}{2}}))e^{-r^2\sin^2{\theta}}(1+\tilde{O}(t^{\frac{1}{2}}))(p(x_0)+\tilde{O}(t^{\frac{1}{2}}))(\int_{ \myT{x_0}{\om_2}}e^{-\frac{\|x_2-y'\|^2}{t}}dy'+O(1))\nonumber\\
=& \frac{1}{\sqrt{t}} C_3 p(x_0) r e^{-r^2\sin^2{\theta}}\partial_{\n_1} f_1 (x_0) + o\left(\frac{1}{\sqrt{t}}\right),
\label{eqn:term1}
\end{align}
where $C_3= \int_{\myT{x_0}{\om_2}}e^{-\|u\|^2} du = \int_{\R^d} e^{-\|u\|^2} du=\pi^{d/2}$.
Note that we use the fact that $\frac{1}{\sqrt{t}}\tilde{O}(\sqrt{t})=O\left(\frac{1}{t^{\varepsilon}}\right)=o\left(\frac{1}{\sqrt{t}}\right)$ in the above derivation. From the first line to the second line we perform a change of variable.
From the third line to the fourth line in the above derivation, we relax $\int_B e^{-\|x_2-y'\|^2/t}dy'$ to be $\int_{ \myT{x_0}{\om_2}}e^{-\|x_2-y'\|^2/t}dy'+O(e^{-t^{-\varepsilon}}) = \int_{ \myT{x_0}{\om_2}}e^{-\|x_2-y'\|^2/t}dy'+O(1)$ based on the result from Appendix B in \cite{CoifmanLafon2006}.

For the second item in Eqn (\ref{eqn:A1}), we have
\begin{align}
& \frac{1}{t}\int_{\B(x)\cap \Om_2}K_t(x,y)(f(x_0)-f(y))p(y)dy \nonumber\\
=& \frac{1}{t} \int_{\B(x)\cap \Om_2}K_t(x,y)(f(x_0)-f(x_2) + f(x_2) - f(y))p(y)dy  \nonumber \\
=&\frac{1}{t}[f(x_0) - f(x_2)] \int_{\B(x)\cap \Om_2}K_t(x,y) dy + \frac{1}{t} \int_{\B(x)\cap \Om_2}K_t(x,y)(f(x_2) - f(y))dy \nonumber \\
=& \frac{1}{\sqrt{t}}C_3 p(x_0) r\cos{\theta} \cdot e^{-r^2 \sin^2{\theta}}\partial_{\n_2} f_2 (x_0)
 + o\left(\frac{1}{\sqrt{t}}\right)
\label{eqn:term2}
\end{align}
Intuitively, the first term in the third line is bounded by the quantity in the last line by a similar argument as the one carried out in Eqn (\ref{eqn:term1}) above. For the second term in the third line, observe that if we replace the kernel $K_t(x,y)$ by $K_t(x_2, y)$, then $\frac{1}{t} \int_{\B(x)\cap \Om_2}K_t(x_2,y)(f(x_2) - f(y))dy$ roughly corresponds to the standard weighted graph Laplacian of function $f_2$ at an interior point $x_2$ in $\Om_2$. Furthermore, by a derivation similarly to the one in Eqn (\ref{eqn:kernelapprox}), replacing $K_t(x,y)$ by $K_t(x_2,y)$ only changes the integral by a factor of $e^{-r^2 \sin^2 \theta}$. Hence the second term can be bounded by $o(\frac{1}{\sqrt{t}})$. See Appendix \ref{appendix:sec:int} for details \footnote{Note $x_2$ is in $\myT{x_0}{\om_2}$. In Eqn (\ref{eqn:term2}), to illustrate the intuition behind the derivation, we abuse the notation slightly and use $f(x_2)$ to refer to $f(\pi^{-1}(x_2))$, where $\pi^{-1}(x_2)$ is the point from $\Om_2$ whose projection in $\myT{x_0}{\om_2}$ is $x_2$.}.
The theorem then follows from  Eqn \ref{eqn:term1} and \ref{eqn:term2}.
{\hfill{\hfill\rule{2mm}{2mm}}}

\yusuremove{
For the second item in Eqn (\ref{eqn:A1}), we have
\begin{align}
& \frac{1}{t}\int_{\B(x)\cap \Om_2}K_t(x,y)(f(x_0)-f(y))p(y)dy \nonumber \\
=& \frac{1}{t^{\frac{d+1}{2}}}e^{-r^2\sin^2{\theta}}(1+\tilde{O}(\sqrt{t}))(p(x_0)+\tilde{O}(\sqrt{t}))\int_{R}e^{-\frac{\|x_2-y'\|^2}{t}}[f(x_0)-f(x_2)+f(x_2)-f(y)] dy'\nonumber \\
=& \frac{1}{t^{\frac{d+1}{2}}}e^{-r^2\sin^2{\theta}}(1+\tilde{O}(\sqrt{t}))(p(x_0)+\tilde{O}(\sqrt{t}))\cdot \nonumber\\
&\left[ (f(x_0)-f(x_2))\int_{\myT{x_0}{\om_2}}e^{-\|x_2-y'\|^2/t}dy'
 + \int_{\myT{x_0}{\om_2}}e^{-\|x_2-y'\|^2/t}(f(x_2)-f(y'))dy' +O(e^{-t^{-\varepsilon}})\right] \nonumber\\
=& \frac{1}{\sqrt{t}}C_3 p(x_0) r\cos{\theta} \cdot e^{-\sin^2{\theta}r^2}\partial_{\n_2} f_2 (x_0)
 + o\left(\frac{1}{\sqrt{t}}\right)
\label{eqn:term2}
\end{align}
The theorem then follows from  Eqn \ref{eqn:term1} and \ref{eqn:term2}.
{\hfill{\hfill\rule{2mm}{2mm}}} \\
}

From the theorem, we can see that for a point $x$ on the intersection or near intersection, $\myL_tf(x)$ is of the form $\frac{1}{\sqrt{t}} C re^{-r^2}$ where the coefficient $C$ is determined by the derivatives of $f$ and the position of $x$.
Furthermore, for a point $x$ on the intersection, we have $r=0$. Hence it follows that the order of $\myL_tf(x)$ \emph{at} an intersection point is {\it the same} as those at \regular{} points, i.e., $O(1)$, instead of order $O(1/\sqrt{t})$ as for points \emph{near} singularities.

\subsection{\Corner: Gluing Boundaries}


We now consider the case where two manifolds are glued along (a part of) their boundaries. The behavior of graph Laplacian for \corner{} points is in some sense the combination of that for \bndS{} and for \intersection{} points: locally, a point can both ``see" the boundary (thus will have terms arised from boundary effect) and the other manifold (which has effects similar to those produced by Eqn (\ref{eqn:A1})). See Appendix \ref{appendix:sec:edge} for the proof of the following theorem.
\begin{theorem}\label{thm:edge}
Let $\Om_i$ ($i=1,2$) be two $d$-dimensional smooth manifolds in $\R^N$ with interior $\om_i$ and nonempty boundary $\d\om_i$. Assume their shared boundary $\d\om_1 \cap \d\om_2$ is a $d-1$-dimensional smooth manifold. Let $f$ be a continuous function over $\Om = \Om_1\cup \Om_2$ whose restriction $f_i:= f|_{\Om_i},i=1,2$ is $C^2$-continuous. Given a point $x\in\om_1$ near the glued boundary, let $x_0$ be its nearest neighbor in $\d\om_1 \cap \d\om_2$, and put $\|x - x_0\| = r\sqrt{t}$. Then for a sufficiently small $t$, we have that
\vspace*{-0.05in}\begin{equation}
\myL_t f(x) =-\frac{1}{\sqrt{t}}p(x_0)[\alpha(r,\theta)
\d_{\n_1}f(x_0)+\beta(r,\theta)\d_{\n_2}f(x_0)]+o\left(\frac{1}{\sqrt{t}}\right),
\end{equation}
where $\theta$ is the angle between the two inward boundary normal $\n_1$ and $\n_2$ of $\d \om_1$ and $\d \om_2$ at $x_0$ respectively, \\
\hspace*{0.1in}$\alpha(r,\theta)=\frac{1}{2}\pi^{(d-1)/2}e^{-r^2}-r\pi^{d/2}\Phi(\sqrt{2}r\cos{\theta})e^{-r^2\sin^2{\theta}}$, \\
\hspace*{0.1in}$\beta(r,\theta)= \frac{1}{2}\pi^{(d-1)/2}e^{-r^2} + r\pi^{d/2}\Phi(\sqrt{2}r\cos{\theta})\cos{\theta}e^{-r^2\sin^2{\theta}}$, \\
and $\Phi(x)$ is the cumulative distribution function of the standard normal distribution.

Furthermore, if $x$ is on the edge, i.e., $r=0$, then we have
\vspace*{-0.05in}\[
\myL_t f(x) =-\frac{1}{2\sqrt{t}}p(x)\pi^{(d-1)/2}[
\d_{\n_1}f(x)+\d_{\n_2}f(x)]+o\left(\frac{1}{\sqrt{t}}\right).
\]
\end{theorem}

The theorem indicates that for a point $x$ near $\d\om_1 \cap \d\om_2$, the limit of $\myL_t f(x)$ is close to the weighted sum of two normal gradients. As $x$ moves to the set $\d\om_1 \cap \d\om_2$, the weights for the two normal gradients will be equal. Finally, for a point $x$ on the boundary $\d\om_1 \cap \d\om_2$, the behavior $\myL_t f(x)$ is equivalent to the addition of the two boundary effects from $\Om_1$ and $\Om_2$ together.

\paragraph{Intersection-type singularity again.}
Notice that for the case of two manifolds $\om_1$ and $\om_2$ intersect at $\om_1 \cap \om_2$ of co-dimension $1$, we can actually regard this scenario as four pieces of manifolds, $\om_1^+,\om_1^-,\om_2^+$ and $\om_2^-$, glued together by $\om_1\cap\om_2$. Here, we use $\om_i^+$ and $\om_i^-$ to denote the two pieces of $\om_i$ but on the different side of $\om_1\cap \om_2$.
The advantage of taking this view is that we allow the functions on manifolds $\om_1$ and $\om_2$ to be only $C^0$-continuous at the intersection $\om_1 \cap \om_2$, instead of $C^2$-continuous as required in Theorem \ref{thm:intersection}. (However, note that Theorem \ref{thm:intersection} also holds for the cases when the co-dimension of the intersection is higher than $1$.)
Using the same argument as in the proof of Theorem \ref{thm:edge} to this case of gluing 4 $d$-manifolds along a common $d-1$-boundary, we obtain the following corollary.

\begin{corollary}\label{thm:intersection1}
Let $\Om_1$ and $\Om_2$ be two $d$-dimensional smooth manifolds in $\R^N$ potentially with boundaries, and their intersection $\om_1 \cap \om_2$ is a smooth manifold of dimension $1$. Let $f$ be a continuous function over $\Om = \Om_1 \cup \Om_2$ whose restriction $f_i: = f | _{\Om_i}$ on $\Om_i$, $i = 1,2$, is $C^2$-continuous at any regular point and $C^0$-continuous at points in $\om_1 \cap \om_2$.
Given a point $x\in \om_1$ near the intersection, let $x_0$ be its nearest neighbor in $\om_1 \cap \om_2$, and $x_1$ (resp. $x_2$) be its projection in the tangent space of $\Om_1$ at $x_0$ (resp. in the tangent space of $\Om_2$ at $x_0$).
For a sufficiently small $t$, we have
\begin{equation}\label{eqn:intNeighbor1}
\myL_t f(x) = -\frac{\pi^{(d-1)/2}}{2}\frac{1}{\sqrt{t}}p(x)[
\d_{\n_1}f_1^+(x_0)+\d_{(-\n_2)}f_1^-(x_0)+\d_{\n_2}f_2^+(x_0)+\d_{(-\n_2)}f_2^-(x_0)]
+o\left(\frac{1}{\sqrt{t}}\right)
\end{equation}
where $\d_{\n_1}f_1^+(x_0),\d_{(-\n_1)}f_1^-(x_0)$ are the directional derivatives of $f_1$ on two sides of $\Om_1\cap \Om_2$, same for $\d_{\n_1}f_2^+(x_0),\d_{\n_1}f_2^-(x_0)$.
\end{corollary}

Interestingly, it appears the eigenfunctions for a singular manifold with intersection-type singularity can be only $C^0$-continuous across the intersection (instead of $C^2$-continuous) within each piece of manifold. Hence the above Corollary will be useful in analysing the eigenfunctions around intersection singularities (or co-dimension 1).

\section{Finite Sample Complexity and Convergence Rate}
\label{sec:convergencerate}


\begin{theorem}\label{thm:complexity}
Let $x_1,\cdots,x_n$,  be i.i.d. random variables in $\R^N$ sampled from a probability distribution on $\Om$ with the intrinsic dimension $d$
with density $p(x)$, $0<a\le p(x)\le b <\infty$, defined on the manifold on $\Om$ with the intrinsic dimension $d$.
Let $f$ be a bounded function, $|f(x)| <M$. Then for any $x \in \Om$
we have
\begin{equation}
    P\left(\left|\myL_{n,t}f(x)-\myL_{t}f(x)\right|>\epsilon\right)\le
    2n\exp{\left(-\frac{nt^{d/2+2}\epsilon^2}
    {2C_v+2C_m\epsilon t/3}\right)}
\end{equation}
where $C_v$ and $C_m$ are constants depending on the manifold.
\end{theorem}
The proof is based on an application of the Bernstein inequality and the union bound and can be found in Appendix \ref{appendix:sec:convergencerate}.

\noindent The immediate corollary is that to get an asymptotic convergence for an arbitrary fixed point (that is an error of the order of $o(1)$), we need to select $t$ so that $nt^{d/2+2}/\log(n)\to \infty$. That is, we can choose $t=({\log(n)}/{n})^{\frac{2}{d+4}}g(n)$, where $g(n)$ is an arbitrary function such that $\lim_{n\to\infty}g(n)=\infty$.

On the other hand, for points near the singular set the scaling of the operator $\myL_{n,t}$ changes. While inside the domain  $\myL_{n,t}f(x) =O(1)$ for a smooth (fixed) function $f$, for $x$ on the singular set or sufficiently close to it (within $\sqrt{t}$ distance), $\myL_{n,t}f(x) =O(\frac{1}{\sqrt{t}})$.
Thus an accurate estimate for the appropriately rescaled operator requires that $nt^{d/2+1}/\log(n)\to \infty$ leading to $t=({\log(n)}/{n})^{\frac{2}{d+2}}g(n)$. This may seem counter-intuitive as fewer points are required on the singularity than inside the domain to obtain an accurate estimate. One explanation is that near a singularity we are effectively  estimating a degree one differential operator, while inside the domain we are estimating the Laplace-Beltrami operator, which is degree two.

%
%
%
%
%


\section{Discussion of Impact on Laplacian Eigenfunctions}
\label{sec:implication}

Eigenfunctions of graph Laplacians obtained from data play an important role in a variety of applications from spectral clustering to dimensionality reduction and semi-supervised learning. While full proof of their convergence is likely to be very subtle and is beyond the scope of this paper (see~\cite{CLEM_08} for the proof of eigenfunction convergence for the smooth case), we would like to discuss the implications of our results for eigenfunctions under the assumption that such convergence takes place.
Let $\phi_t$ be an eigenfunction of $\myL_t$ (normalized to norm $1$ in $L_2$). We put
$\lambda = \lim_{t \to 0} \lambda_t $ and  $\phi = \lim_{t \to 0} \phi_t$ (with the functions converging in $L_2$ norm). We will also  assume that the necessary derivatives exist.

Below, we briefly discuss the impact of each type of singularity. Some numerical experiments are provided in Appendix \ref{appendix:sec:eigenexp}. For simplicity, we will assume the samples to be infinite and the multiplicity of each eigenfunction to be one.


\paragraph{Boundary singularity:}
For a point $x$ on or near (within $\sqrt{t}$ of) the boundary our results indicate
$$\lambda_t \phi_t (x) = \myL_t \phi_t (x) = C  \frac{1}{\sqrt{t}} \partial_\n
\phi(x)+o\left(\frac{1}{\sqrt{t}}\right) $$
where $C$ is a constant independent of $t$. Thus, for small values of $t$, and
$\lambda_t \phi_t(x) = O(\frac{1}{\sqrt{t}})\partial_\n \phi_(x)$, which is clearly impossible unless $\partial_\n \phi_t(x) = O(\sqrt{t})$.
Passing to the limit\footnote{Since this is an informal argument, the subtleties related to convergence in $L_2$ are ignored.}
we see  that for $x$ on the boundary, $\partial_\n \phi(x)=0$, meaning that the limit eigenfunctions of the graph Laplacian automatically satisfy  the Neumann boundary conditions. The experimental results in Appendix \ref{appendix:sec:eigenexp} are consistent with this finding.


\paragraph{Edge singularity:}
Consider now a point $x$ at the edge where two "sheets" $\om_1$ and $\om_2$ come together.
From Theorem~\ref{thm:edge} we have
\[
\myL_t \phi_t(x) =\frac{C}{\sqrt{t}}(
\d_{\n_1}\phi_t(x)+\d_{\n_2}\phi_t(x)]+o\left(\frac{1}{\sqrt{t}}\right).
\] where $\n_1$ and $\n_2$ are the two directions normal to the singular set at $x$.
By an argument following the one above, we have
$
   \d_{\n_1} \varphi_t(x)+\d_{\n_2} \varphi_t(x)=O(\sqrt{t}).
$ and hence $$\d_{\n_1} \varphi(x)+\d_{\n_2} \varphi(x)= 0$$
In other words the directional derivatives for the limit eigenfunction must cancel each other.
This is quite surprising for the following reason: suppose, that the edge singularity is obtained by folding a smooth manifold (imagine folding a sheet of paper, creating an edge). Then the above condition means that the eigenfunctions of the original surface are invariant under this "folding", despite the fact that the graph Laplacian operator is very different near the
singular set! In other words if a singular surface is isometric to a smooth manifolds, the spectral structure appears to be preserved, even though the distances computed for points near the edge are very far from the intrinsic distances on smooth and
the operator near the edge is not the Laplace-Beltrami operator.

Some numerical  illustrations of this phenomenon are given in the Appendix \ref{appendix:sec:eigenexp}.

\paragraph{Intersection singularity:} For the intersection singularity the analysis follows that of the edge case, but the implications for the shape of eigenfunctions are not completely clear, and warrant further investigation. One intriguing observation is that if the co-dimension of the singularity is greater than one, it has  only a local effect on the shape of eigenfunctions. The situation is analogous to the disjoint union of manifolds. The reason is easy to see -- the volume of $\sqrt{t}$-neighborhood $B$ around the singular set of co-dimension at least two, will be at most  $O((\sqrt{t})^2) = O(t)$. Recall that an eigenfunction minimizes
the quantity $\langle L_t \phi_t, \phi_t \rangle_{L_2(p)}$ under orthogonality conditions. We see that the contribution of the points around the singular set is bounded by $O(t) O(\frac{1}{\sqrt{t}}) = O(\sqrt{t})$ and vanishes as $t\to0$.

\newpage


\newpage
\appendix

\section{Proofs for Theorems on Graph Laplacian Limit}
\label{appendix:sec:infinite}

\subsection{Sketch of Proof for Theorem 1} \label{appendix:sec:bnd}

For a sufficiently small $t$, let $\Omega_{bd}$ be the set of points
that are within distance $t^{1/2 - \epsilon}$ from the boundary $\d\Omega$ (a
thin layer of ``shell''), where $\epsilon$ is any sufficiently small positive constant.
Set $\Omega_{in}={\Om}/\Omega_{bd}$.
We will show that for a small $t$, $\myL_{t}f(x)$ is approximated by
two different terms on $\Omega_{db}$ and $\Omega_{in}$, and more
importantly they have different orders of $t$.

\paragraph{Points away from boundary.}
Given a point $x\in \om$, let $\B(x)$ denote the ball of radius $t^{1/2 - \epsilon}$ centered at $x$, where $\epsilon$ is any sufficiently small positive constant.
First, we approximate the integral in Eqn (\ref{eqn:aimintegral}) when the integral is taken inside the ball
$\B(x)$. By results from Appendix B of \cite{CoifmanLafon2006}, the error induced by constraining the integral to within $\B(x) \cap \Om$ is only $O(e^{-t^{-\epsilon}})$.

Now given a point $y\in\B(x)$, let $u$ be the projection of $y$ onto the tangent space $\myT{x}{\om}$ of $\Om$ at $x$. Choose $x$ to be the origin of $\myT{x}{\om}$. We have the following approximations for each of the three terms in the integral in Eqn (\ref{eqn:aimintegral}) (see e.g, \cite[Chapter
4.2]{belkinThesis} and \cite[Appendix B]{CoifmanLafon2006}). Here $K( a)$ denotes $K(a) = e^{-\frac{a}{t}}$. $N$ and $d$ are the dimension of the ambient space and of the tangent space, respectively.
\begin{equation}\label{equ:detailedApprox}
\begin{array}{rl}
    K(\|x-y\|^2_{\R^N})= & K(\|u\|^2_{\R^d})+O(\|u\|^4_{\R^d})\\
    \\
    f(x)-f(y)= &-u^T \nabla f(x) - \frac{1}{2}u^TH(x)u + O(\|u\|^{3}_{\R^d})\\
    \\
    p(y)=&p(x)+u^T\nabla p(x)+O(\|u\|^{2}_{\R^d})\\
\end{array}
\end{equation}
where $H(x)$ is the Hessian of $f(x)$ at $x$.

Now for simplicity let $\tilde{O}(t^{\beta})$ denote $O(t^{\beta - \epsilon})$ for any sufficiently small positive constant $\epsilon > 0$. Let $\pi_x: \Om \rightarrow \myT{x}{\om}$ be the projection from $\Om$ onto the tangent space $\myT{x}{\om}$, and set $R := \pi_x(\B(x) \cap \Om)$. Obviously, $R \subset \myT{x}{\om}$.
Combing the above approximations together with the fact that $y \in \B(x)$ and thus $\| x - y \| = \tilde{O}(t)$, and using a change of
variable $u\to \sqrt{t}w$, we obtain the following:

\begin{equation}\label{equ:bd:decomposition}
\begin{array}{rl}
    \myL_{t}f(x)=&\frac{1}{t} \int_{\Om} K_t(x,y)(f(x)-f(y))p(y)dy\\
\\
    =&\frac{1}{t} \int_{\B(x) \cap \Om} K_t(x,y)(f(x)-f(y))p(y)dy+ O(e^{-t^{-\epsilon}})\\
 \\
   =&-\frac{1}{t^{d/2}}\int_{R} \frac{1}{t}K(\|u\|^2_{\R^d})[
    (\sqrt{t}u^T \nabla f(x)+\frac{t}{2} u^T H(x) u )\times\\
    &  \qquad \qquad \qquad \qquad\ \ (p(x)+\sqrt{t}u^T\nabla p(x))]t^{d/2}J \pi^{-1}|_{u}du + \frac{1}{t} \cdot \tilde{O}(t^{1/2})\\
    =&-\int_{\myT{x}{\om}} e^{-\|w\|^2_{\R^d}} \{\frac{1}{\sqrt{t}}[p(x)(w^T\nabla f(x))]+\\
    & \qquad \qquad \qquad \qquad \quad [w^T \nabla f(x)\times w^T \nabla
    p(x)+\frac{1}{2}p(x)w^TH(x)w]\} dw + o(\frac{1}{\sqrt{t}})
\end{array}
\end{equation}
From the first line to the second line, we replace the integral
over $\Om$ with ball $\B(x)$, generating an exponentially small
error for sufficiently small $t$ \cite[Appendix B]{CoifmanLafon2006}.
From the second line to the third line, by changing the variable from $y$ to $\pi_x(y) = u$,
this integral can be rewritten as an integral over the region $R \subset \myT{x}{\om}$.
In the fourth line, we relax this integral over $R$ to be the integral over the entire tangent space $\myT{x}{\om}$,
plus another exponentially small error $O(e^{-t^{-\epsilon}})$ which is consumed by $o(\frac{1}{\sqrt{t}})$.

For an interior point $x \in \om_{in}$, $\myT{x}{\om}$ is a $d$-dimensional linear space, and the function $e^{-\|w \|^2_{\R^d}}$ is an even function on $\myT{x}{\om}$.
When taking the integral, the first term in the last integral in Eqn (\ref{equ:bd:decomposition}) with order
$O\left(\frac{1}{\sqrt{t}}\right)$ is odd and therefore vanishes.
The remaining three terms in the last line of Eqn (\ref{equ:bd:decomposition})
are of order $O(1)$ inside the integral, and they are
exactly the weighted Laplacian at $x$ as previously studied in the literature, and hence
$\myL_t f(x) = o(\frac{1}{\sqrt{t}})$.
The kernel is however not symmetric for points on or near the boundary, which are studied next.

\begin{figure}[h]
\begin{center}
\includegraphics[width=0.5\columnwidth]{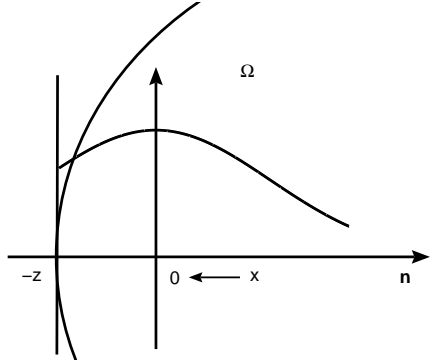}
\caption{Gaussian weight at $x$ near the
boundary.}\label{fig:bd:integral}
\end{center}
\end{figure}

\paragraph{Points near boundary.}
We now consider $x\in \om_{bd}$ (the ``shell'') near the boundary.
See Figure~\ref{fig:bd:integral}, where we show the local neighborhood around $x$ projected in $\myT{x}{\om}$.
Let $x_0$ be the  nearest boundary point on $\partial \om$ to $x$.
Choose a local coordinate system of $\myT{x}{\om}$ with $x$ at the origin, and assume the projection of $x_0$ in $\myT{x}{\om}$ is at $-z$. Let $\n$ the unit direction along $x-z$.
When $x$ is close to the boundary, the kernel $K_t(x,y)$ is no longer symmetric. In particular, consider an orthonormal coordinate system around $x$, the Gaussian kernel is symmetric in all coordinate-axis other than along $x-z$.
Along the direction $\n$, the Gaussian convolution is from $-z / \sqrt{t}$ (as we have changed variable from $u \rightarrow \sqrt{t}w$) to $+\infty$,
which is not symmetric.
Therefore, $e^{-\|w\|^2_{\R^d}}$ is not an
even function in the normal direction $\n$, and the integral of the highest order term (i.e, the first term of $O\left(\frac{1}{\sqrt{t}}\right)$) in the last integral of
Eqn (\ref{equ:bd:decomposition}) will not vanish.

Specifically, let $u = (u_1, u_2, \ldots, u_d)$ be the coordinate of $u$ in the aforementioned coordinate system of $\myT{x}{\om}$. Assume that $u_1$ is the coordinate along the normal direction $\n$. Recall that we have applied the change of variable $u \rightarrow \sqrt{t} w$. So let $w = (w_1, w_2, \ldots, w_d)$ be the coordinate of $w$.
Since we have assumed in the theorem that $\| x- x_0 \| = r\sqrt{t}$, we have that $\frac{z}{\sqrt{t}} = r + O(r^3)$.
When we integrate the last integral in Eqn (\ref{equ:bd:decomposition}), all the odd terms of $u_i$ still vanish in all
directions except the normal direction $\n$, and the most important
point is that the leading term along the normal direction is of
order $O\left(\frac{1}{\sqrt{t}}\right)$, different from $O(1)$ for the
interior points. In particular, we have: 
\begin{align*}
    \myL_{t}f(x)&=-\frac{1}{\sqrt{t}}p(x)\d_\n f(x)\int_{-\infty}^{+\infty} \cdots \int_{-\infty}^{+\infty}\int_{-z/\sqrt{t}}^\infty
    e^{-\|w\|^2_{\R^d}} w_1 dw_1dw_2\cdots dw_d + o\left(\frac{1}{\sqrt{t}}\right) \\
&= -\frac{1}{\sqrt{t}} \cdot \frac{1}{2}\pi^{\frac{d-1}{2}}e^{-r^2} \partial_{\n} f(x) + o(\frac{1}{\sqrt{t}}).
\end{align*}
The theorem follows from further performing the Taylor expansion of $f(x)$ at $x_0$.


\subsection{Details for the Intersection Case}
\label{appendix:sec:int}

Our goal is to bound the integral $ \frac{1}{t}\int_{\B(x)\cap \Om_2}K_t(x,y)(f(x_0)-f(y))p(y)dy$ as in Eqn (\ref{eqn:term2}).
First, recall that $\pi_{x_0}: \Om_2 \rightarrow \myT{x_0}{\om_2}$ is the projection map from $\Om_2$ to the tangent space of $\Om_2$ at $x_0$.
For points  on $\Om_2$ sufficiently close to $x_0$, this map is in fact a bijection and its inverse $\pi^{-1}$ exsits.
Hence for $t$ sufficiently small (and thus $x_2$ is close enough to $x$), $\pi^{-1}(x_2)$ exists and we set it to be $x_3 := \pi^{-1}(x_2) \in \Om_2$.
Since $\|x_0 - x_2 \| = r \cos \theta \sqrt{t} + O(t)$, we have $\| x_2 - x_3 \| = O(t)$ by Eqn (\ref{equ:basicApprox1}).
It follows from triangle inequality that $\| x - x_3 \| = \| x - x_2 \| + O(t) = r \sin \theta \sqrt{t} + O(t)$.
We can then apply a similar derivation as in Eqn (\ref{eqn:kernelapprox}) to show that
\begin{equation}
K_t(x, y) = e^{-r^2 \sin^2 \theta} K_t(x_3, y) \cdot (1+\tilde{O}(\sqrt{t})); ~~\text{and}~~K_t(x,y) =  e^{-r^2 \sin^2 \theta} K_t(x_3, y') \cdot (1+\tilde{O}(\sqrt{t})), \label{eqn:kernelapprox2}
\end{equation}
where $y'$ is the projection of $y$ on $\myT{x_0}{\om_2}$ as defined in Section \ref{sec:inter}.
We now have
\begin{align}
& \frac{1}{t}\int_{\B(x)\cap \Om_2}K_t(x,y)(f(x_0)-f(y))p(y)dy \nonumber \\
=& \frac{1}{t}(f(x_0) - f(x_3)) \int_{\B(x)\cap \Om_2}K_t(x,y)p(y)dy +  \frac{1}{t}\int_{\B(x)\cap \Om_2}K_t(x,y)(f(x_3)-f(y))p(y)dy
\label{eqn:newterm2}
\end{align}
Following the same derivation as in Eqn (\ref{eqn:term1}), and combining with $\| x_0 - x_2 \| = r \cos \theta \sqrt{t} + O(t)$,
the first term can be bound by:
\begin{align}
& \frac{1}{t}(f(x_0) - f(x_3)) \int_{\B(x)\cap \Om_2}K_t(x,y)p(y)dy \nonumber \\
=& \frac{1}{\sqrt{t}} (\| x_0 - x_2 \| \partial_{\n_2} f_2 (x_0) + \tilde{O}(t^{\frac{1}{2}})) e^{-r^2 \sin^2 \theta} (1+\tilde{O}(t^{\frac{1}{2}})) (p(x_0) + \tilde{O}(t^{\frac{1}{2}})) \int_{\myT{x_0}{\om_2}} K_t(x_2, y') dy' \nonumber \\
=& \frac{1}{\sqrt{t}} C_3 p(x_0) r \cos \theta \cdot e^{-r^2 \sin^2 \theta} \partial_{\n_2} f_2(x_0) + o(\frac{1}{\sqrt{t}}).
\label{eqn:newterm2-1}
\end{align}

By using Eqn (\ref{eqn:kernelapprox2}), the second term in Eqn (\ref{eqn:newterm2}) is:
\begin{align}
& \frac{1}{t}\int_{\B(x)\cap \Om_2}K_t(x,y)(f(x_3)-f(y))p(y)dy \nonumber \\
= &  \frac{1}{t}\int_{\B(x)\cap \Om_2} e^{-r^2 \sin^2 \theta} (1+\tilde{O}(\sqrt{t})) \cdot K_t(x_3,y)(f(x_3)-f(y))p(y)dy \nonumber \\
= &e^{-r^2 \sin^2 \theta} (1+\tilde{O}(\sqrt{t})) \cdot \frac{1}{t} \int_{\B(x)\cap \Om_2}K_t(x_3,y)(f(x_3)-f(y))p(y)dy + o(\frac{1}{\sqrt{t}}). \label{eqn:newterm2-2}
\end{align}
Now observe that since the radius of $\B(x)$ is $\Theta(t^{\frac{1}{2} - \varepsilon})$ for a sufficiently small positive $\varepsilon$, and since the distance from $x$ to $x_3$ is $O(t^{1/2})$, we have that there exists some ball $\B'(x_3)$ of radius still in the asymptotic order of $\Theta(t^{\frac{1}{2} - \varepsilon})$ around $x_3$ such that $\B'(x_3)$ is contained inside $\B(x)$. This implies that
$\B'(x_3) \cap \Om_2 \subseteq \B(x) \cap \Om_2$.
Hence by results from Appendix B of \cite{CoifmanLafon2006},
$$\int_ {\B'(x_3) \cap \Om_2} K_t(x_3,y)(f(x_3)-f(y))p(y)dy = \int_ {\B(x) \cap \Om_2} K_t(x_3,y)(f(x_3)-f(y))p(y)dy + O(e^{-t^{-\varepsilon}}).
$$
On the other hand, $\frac{1}{t} \int_{\B'(x_3) \cap \Om_2} K_t(x_3,y)(f_2(x_3)-f_2(y))p(y)dy$ is simply an approximation of the functional Laplacian $\myL_t f_2 (x_3)$ of the function $f_2$ at an interior point $x_3$ over a manifold $\Om_2$. Hence by Eqn (\ref{eqn:interior}),
$$\frac{1}{t} \int_ {\B'(x_3) \cap \Om_2} K_t(x_3,y)(f(x_3)-f(y))p(y)dy = - \frac{1}{2} \pi^{d/2} p(x_3) \Delta_{p^2} f_2(x_3) + o(1) . $$
It then follows from Eqn (\ref{eqn:newterm2-2}) that
\begin{align}
 &\frac{1}{t}\int_{\B(x)\cap \Om_2}K_t(x,y)(f(x_3)-f(y))p(y)dy \nonumber \\
=&  e^{-r^2 \sin^2 \theta} \frac{1}{t} \int_{\B(x)\cap \Om_2}K_t(x_3,y)(f(x_3)-f(y))p(y)dy + o(\frac{1}{\sqrt{t}}) \nonumber \\
=&  - \frac{1}{2} \pi^{d/2} p(x_3) \Delta_{p^2} f_2(x_3) + o(\frac{1}{\sqrt{t}}) = o(\frac{1}{\sqrt{t}}). \label{eqn:newterm2-3}
\end{align}
Putting Eqn (\ref{eqn:newterm2}), (\ref{eqn:newterm2-1}) and (\ref{eqn:newterm2-3}) together, we conclude that
$$\frac{1}{t}\int_{\B(x)\cap \Om_2}K_t(x,y)(f(x_0)-f(y))p(y)dy = \frac{1}{\sqrt{t}} C_3 p(x_0) r \cos \theta \cdot e^{-r^2 \sin^2 \theta} \partial_{\n_2} f_2(x_0) + o(\frac{1}{\sqrt{t}}), $$
which is what we claimed in Eqn (\ref{eqn:term2}).

\subsection{Proof for \Corner{} Points}
\label{appendix:sec:edge}

\begin{figure}[ht]
\begin{center}
\includegraphics[width=0.5\columnwidth]{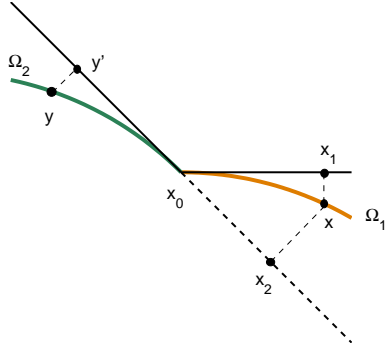}
\caption{Edge case.}\label{fig:lcorner}
\end{center}
\end{figure}
Suppose we have two manifolds forming the ``edge'', denote them by $\Om_1$ and $\Om_2$. We use the same notation as in the proof for the intersection case; see  the illustration in Figure \ref{fig:lcorner}. In particular, let $x_0$ denote the closest point from $\partial \Om_1 \cap \partial \Om_2$ to a point $x \in \om_1$. Let $x_1$ and $x_2$ be the projection of $x$ on the tangent space $\myT{x_0}{\om_1}$ and $\myT{x_0}{\om_2}$ at $x_0$, respectively.
Note that $\|x-x_0\|=r\sqrt{t}$.
Let $f_1,f_2$ be the restriction of $f$ on $\om_1$ and $\om_2$ respectively; we know that $f_1(x_0)=f_2(x_0)$. Again, $\B(x)$ denotes the ball centered at $x$ with radius $t^{\frac{1}{2}-\varepsilon}$. Then we have
$$\frac{1}{t}\int_{\Om} K_t(x,y)(f(x)-f(y))p(y)dy = \frac{1}{t}\int_{\B(x)\cap \Om} K_t(x,y)(f(x)-f(y))p(y)d + o(1); $$
and
\begin{align}
 & \frac{1}{t}\int_{\B(x)\cap \Om} K_t(x,y)(f(x)-f(y))p(y)dy  \nonumber\\
=& \frac{1}{t}\int_{\B(x)\cap \Om_1} K_t(x,y)(f_1(x)-f_1(y))p(y)dy+
\frac{1}{t}\int_{\B(x)\cap \Om_2}
K_t(x,y)(f_1(x)-f_2(y))p(y)dy
\label{eqn:C}
\end{align}
The first integral in Eqn (\ref{eqn:C}) on $\B(x)\cap \Om_1$ is exactly what we had before for the boundary singularity (recall the second line in Eqn (\ref{equ:bd:decomposition})), which we have shown to be the following in Appendix \ref{appendix:sec:bnd}:
\begin{equation}\label{eqn:int1}
-\frac{1}{\sqrt{t}}\cdot \frac{1}{2}{\pi}^{(d-1)/2}p_1(x_0)e^{-r^2}\langle \nabla f_1, \n_1\rangle+o\left(\frac{1}{\sqrt{t}}\right).
\end{equation}

Now consider the second integral in Eqn (\ref{eqn:C}) on $\B(x)\cap \Om_2$:
\begin{align}
 & \frac{1}{t}\int_{\B(x)\cap \Om_2}K_t(x,y)(f_1(x)-f_2(y))p(y)dy \nonumber\\
=& \frac{1}{t}\int_{\B(x)\cap \Om_2}K_t(x,y)(f_1(x)-f_2(x_0)+f_2(x_0)-f_2(y))p(y)dy \nonumber \\
=& \frac{1}{t}\int_{\B(x)\cap \Om_2}K_t(x,y)(f_1(x)-f_1(x_0))p(y)dy+\frac{1}{t}\int_{\B(x)\cap \Om_2}K_t(x,y)(f_2(x_0)-f_2(y))p(y)dy \nonumber \\
=&
\frac{1}{t}(f_1(x)-f_1(x_0))\int_{\B(x)\cap \Om_2}K_t(x,y)p(y)dy+\frac{1}{t}\int_{\B(x)\cap \Om_2}K_t(x,y)(f_2(x_0)-f_2(y))p(y)dy
\label{eqn:D}
\end{align}

Let $\pi_x$ denote the projection from $\Om_2$ to $\myT{x_0}{\om_2}$, and set $y' = \pi_x(y)$ for any point $y \in \Om_2$.
Since the notation is the same with intersection case, we have by Eqn (\ref{eqn:kernelapprox})
\[
K_t(x,y)=\frac{1}{t^{d/2}}(e^{-r^2 \sin^2{\theta}}e^{-\|x_2-y'\|^2/t}(1+\tilde{O}(t^{1/2}))).
\]

Set $B:= \pi_x(\B(x) \cap \Om_2)$ as the image of $\B(x) \cap \Om_2$ in the tangent space $\myT{x_0}{\om_2}$.
Let $\myT{x_0}{\om_2}^+$ denote the half-space of the tangent space $\myT{x_0}{\om_2}$ that contains $R$.
Combing the bounds in Eqn (\ref{equ:basicApprox1}) and (\ref{eqn:jacobianapprox}), for the first integral in Eqn (\ref{eqn:D}), we have
\begin{align}
&\frac{1}{t}(f_1(x)-f_1(x_0))\int_{\B(x)\cap \Om_2}K_t(x,y)p(y)dy ~~=~~ \frac{1}{t}(f_1(x)-f_1(x_0))\int_{B}K_t(x,y)p(y) J \pi_x^{-1} |_{y'} dy' \nonumber\\
=& \frac{1}{t^{\frac{d}{2}+1}}(f_1(x)-f_1(x_0))e^{-r^2 \sin^2{\theta}}\int_{B}e^{-\|x_2-y'\|^2/t}(1+\tilde{O}(t^{1/2}))(p(x_0)+\tilde{O}(t^{1/2})) dy' \nonumber \\
=& \frac{1}{t^{\frac{d}{2}+1}}(f_1(x)-f_1(x_0))e^{-r^2 \sin^2{\theta}}(1+\tilde{O}(t^{1/2}))(p(x_0)+\tilde{O}(t^{1/2}))\left(\int_{\myT{x_0}{\om_2}^+}e^{-\|x_2-y'\|^2/t}dy' +O(e^{-t^{-\varepsilon}})\right)\nonumber \\
=& C(r,\theta)\frac{1}{\sqrt{t}}p_2(x_0)r e^{-r^2 \sin^2{\theta}}\langle\nabla f_1 (x_0),
\n_1\rangle+o\left(\frac{1}{\sqrt{t}}\right),  \label{eqn:int2}
\end{align}
where
$C(r,\theta)=\int_{\myT{x_0}{\om_2}^+}e^{-\|x_2-y'\|^2/t}dy' = \int_{\R^d,y_d>-r\cos(\theta)}{e^{-\|y\|^2}dy}=\pi^{d/2}\Phi(\sqrt{2}r\cos{\theta})$,
and $y_d$ is the $d$-th entry of coordinate of $y$ in the $d$-dimensional tangent space.
The last step above also involves applying the following formula:
$$f_1(x) - f_1(x_0) = \langle \nabla f_1 (x_0), x_1 - x_0 \rangle + \tilde{O}( \| x - x_0 \|^2) = r \sqrt{t} \langle \nabla f_1 (x_0), \n_1 \rangle + \tilde{O}(t).$$

Now for the second integral in Eqn (\ref{eqn:D}), we have:
\begin{align}
& \frac{1}{t}\int_{\B(x)\cap \Om_2}K_t(x,y)(f_2(x_0)-f_2(y))p(y)dy \nonumber\\
=& \frac{1}{t}\int_{\B(x)\cap \Om_2}K_t(x,y)\langle \nabla f_2\big | _{x_0}, x_0-y\rangle p(y)dy \nonumber \\
=& \frac{1}{t}\int_{\B(x)\cap \Om_2}K_t(x,y)\langle \nabla f_2\big | _{x_0}, x_0-x_2\rangle p(y)dy + \frac{1}{t}\int_{\B(x)\cap \Om_2}K_t(x,y)\langle \nabla f_2\big | _{x_0}, x_2-y\rangle p(y)dy \label{eqn:yusuA}
\end{align}
The second term from the above equation is exactly the same as if we were considering a point $x_2$ nears the boundary of $\Om_2$. Hence using the results from Appendix \ref{appendix:sec:bnd}, we have that
\begin{align}
& \frac{1}{t}\int_{\B(x)\cap \Om_2}K_t(x,y)\langle \nabla f_2\big | _{x_0}, x_2-y\rangle p(y)dy \nonumber \\
=&\frac{1}{t^{\frac{d}{2}+1}}e^{-r^2\sin^2\theta}\int_{B}e^{-\|x_2-y'\|^2/t}\langle \nabla f_2\big | _{x_0}, x_2-y\rangle (p(x_0)+\tilde{O}(t^{1/2}))dy \nonumber \\
=&\frac{1}{t^{\frac{d}{2}+1}}e^{-r^2\sin^2\theta}(p(x_0)+\tilde{O}(t^{1/2}))\langle \nabla f_2\big | _{x_0},\int_{B}e^{-\|x_2-y'\|^2/t} (x_2-y )dy \rangle \nonumber \\
=&\frac{1}{t^{\frac{d}{2}+1}}e^{-r^2\sin^2\theta}(p(x_0)+\tilde{O}(t^{1/2}))\langle \nabla f_2\big | _{x_0},\int_{\myT{x_0}{\om_2}^+}e^{-\|x_2-y'\|^2/t} (x_2-y) dy \rangle \nonumber \\
=&\frac{1}{t^{\frac{d}{2}+1}}e^{-r^2\sin^2\theta}(p(x_0)+\tilde{O}(t^{1/2}))(- \frac{t^{d/2}}{2}\pi^{\frac{d-1}{2}}e^{-r^2\cos^2\theta} \partial_{\n_2} f_2) \nonumber \\
= &- \frac{1}{\sqrt{t}} \cdot \frac{1}{2} \pi^{\frac{d-1}{2}} e^{-r^2} \partial_{\n_2} f_2 (x_0) + o(\frac{1}{\sqrt{t}}).
\label{eqn:yusuB}
\end{align}


\noindent For the first term from Eqn (\ref{eqn:yusuA}), we have:
\begin{align}
& \frac{1}{t}\int_{\B(x)\cap \Om_2}K_t(x,y)\langle \nabla f_2\big | _{x_0}, x_0-x_2\rangle p(y)dy \nonumber \\
=& \frac{1}{t^{\frac{d}{2}+1}}e^{-r^2 \sin^2{\theta}} \int_{B}e^{-\|x_2-y'\|^2/t}(1+\tilde{O}(t^{1/2}))\langle\nabla f_2(x_0), x_0-x_2\rangle(p(x_0)+\tilde{O}(t^{1/2}))dy' \nonumber \\
=& \frac{1}{t^{\frac{d}{2}+1}}e^{-r^2 \sin^2{\theta}}(1+\tilde{O}(t^{1/2}))(p(x_0)+\tilde{O}(\sqrt{t}))\int_{B}e^{-\|x_2-y'\|^2/t}\langle\nabla f_2(x_0), x_0-x_2\rangle dy'\nonumber\\
=& \frac{1}{t^{\frac{d}{2}+1}}e^{-r^2 \sin^2{\theta}}p(x_0)\langle \nabla f_2(x_0), x_0-x_2\rangle \int_{R}e^{-\|x_2-y'\|^2/t}dy' \nonumber\\
=& \frac{1}{t^{\frac{d}{2}+1}}e^{-r^2 \sin^2{\theta}}p(x_0)\langle \nabla f_2, x_0-x_2\rangle \left(\int_{ \myT{x_0}{\om_2}^+}e^{-\|x_2-y'\|^2/t}dy'+O(e^{-t^{-\varepsilon}})\right) \nonumber\\
= &\frac{1}{\sqrt{t}}p(x_0) C(r,\theta) \cdot r \cos{\theta}e^{-r^2 \sin^2{\theta}}\langle\nabla f_2(x_0), \n_2\rangle+o\left(\frac{1}{\sqrt{t}}\right), \label{eqn:yusuC}
\end{align}
where $C(r, \theta)$ is defined earlier.
Let $x'$ be the projection of $x_1$ on the tangent space $\myT{x_0}{\om_2}$.
The last step in Eqn (\ref{eqn:yusuC}) uses the fact that $\| x_0 - x_2 \| = r \sqrt{t} \cdot | \cos \theta| + O(t)$.
Combining all the results in (\ref{eqn:int1}), (\ref{eqn:int2}), (\ref{eqn:yusuA}), (\ref{eqn:yusuB}), and
(\ref{eqn:yusuC}), we prove Theorem \ref{thm:edge}.

\yusuremove{
Now for the second integral in Eqn (\ref{eqn:D}), we have
\begin{align}
 & \frac{1}{t}\int_{\B(x)\cap \Om_2}K_t(x,y)(f_2(x_0)-f_2(y))p(y)dy \nonumber\\
=& \frac{1}{t^{\frac{d}{2}+1}}e^{-r^2 \sin^2{\theta}}\int_{R}e^{-\|x_2-y'\|^2/t}(1+\tilde{O}(t^{1/2}))\langle\nabla f_2(x_0), x_0-y'\rangle(p(x_0)+\tilde{O}(t^{1/2}))dy' \nonumber \\
=& \frac{1}{t^{\frac{d}{2}+1}}e^{-r^2 \sin^2{\theta}}(1+\tilde{O}(t^{1/2}))(p(x_0)+\tilde{O}(\sqrt{t}))\int_{R}e^{-\|x_2-y'\|^2/t}\langle\nabla f_2(x_0), x_0-x_2+y_0-y'\rangle dy'\nonumber\\
=& \frac{1}{t^{\frac{d}{2}+1}}e^{-r^2 \sin^2{\theta}}p(x_0)\langle \nabla f_2(x_0), x_0-x_2\rangle \int_{R}e^{-\|x_2-y'\|^2/t}dy' \nonumber\\
 &+\frac{1}{t^{\frac{d}{2}+1}}e^{-r^2 \sin^2{\theta}}p(x_0)\langle \nabla f_2(x_0), \int_{R}e^{-\|x_2-y'\|^2/t}(x_2-y')dy'\rangle+o\left(\frac{1}{\sqrt{t}}\right)\nonumber\\
=& \frac{1}{t^{\frac{d}{2}+1}}e^{-r^2 \sin^2{\theta}}p(x_0)\langle \nabla f_2, x_0-x_2\rangle \left(\int_{ \myT{x_0}{\om_2}^+}e^{-\|x_2-y'\|^2/t}dy'+O(e^{-t^{-\varepsilon}})\right) \nonumber\\
 &+\frac{1}{t^{\frac{d}{2}+1}}e^{-r^2 \sin^2{\theta}}p(x_0)\langle \nabla f_2, \left(\int_{ \myT{x_0}{\om_2}^+}e^{-\|x_2-y'\|^2/t}(x_2-y')dy'+O(e^{-t^{-\varepsilon}})\right)\rangle+o\left(\frac{1}{\sqrt{t}}\right)  \label{eqn:intermediate} 
\end{align}
Now using Eqn (\ref{equ:basicApprox1}), we can show that $\| x_0 - x_2 \| = r \sqrt{t} |\cos \theta| + O(t)$. Hence the first term in the last line of above Equation equals
$$\frac{1}{\sqrt{t}}p(x_0) C(r,\theta) \cdot r \cos{\theta}e^{-r^2 \sin^2{\theta}}\langle\nabla f_2(x_0), \n_2\rangle+o\left(\frac{1}{\sqrt{t}}\right), $$
where $C(r, \theta)$ is defined earlier.

For the second term in the last line in Eqn (\ref{eqn:intermediate},
Combining all the results in (\ref{eqn:int1}), (\ref{eqn:int2}) and
(\ref{eqn:int3}), we conclude the proof.
}

\section{Proof for Convergence Rate}
\label{appendix:sec:convergencerate}

\begin{lemma}
\label{lemma:convergencerate}
Let $X_1,\cdots,X_n$, Z be i.i.d. random variables of $\R^N$ from density $p(x)\in C^\infty(\Om)$, $0<a\le p(x)\le b <\infty$, $K_t(x,y)$ be a Gaussian weight function and $|f| <M$, then for any $x\in \Om$
\begin{equation}
    P\left(\left|\frac{1}{n}\sum_{i=1}^n
    K_t(x,X_i)f(X_i)-\E_Z[K_t(x,Z)f(Z)]\right|>\epsilon\right)\le 2\exp{\left(-\frac{nt^{d/2}\epsilon^2}
    {2C_v+2C_m\epsilon/3}\right)}
\end{equation}
where $C_v$ and $C_m$ only depend on $d$ and $K(\cdot,\cdot)$, $f(\cdot)$, $p(\cdot)$.
\end{lemma}\label{lemma}
\begin{proof}
Let $W_i(x)=K_t(x,X_i)f(X_i)$, and $Y_i(x)=W_i(x)-\E_{X_i}[W_i(x)]$.
We have $\E_{X_i}[Y_i(x)]=0$, and
\begin{equation}
\begin{array}{rl}
    |Y_i(x)|\le & |W_i(x)|+|\E_{X_i}[W_i(x)]|\\
    =& |K_t(x,X_i)f(X_i)|+|\int_{\Om} K_t(x,y)f(y)p(y)dy|\\
    \le & \frac{1}{t^{d/2}} M + Mb \int_{\Om} K_t(x,y)dy\\
    \le & \frac{M}{t^{d/2}}+Mb C_g=(C_m+Mb C_g t^{d/2})/t^{d/2}
\end{array}
\end{equation}
where $C_g=\int_{\Om}K_t(x,y)dy < \infty$, and $C_m=M$. For
Var$[Y_i(x)]$,
\begin{equation}
\begin{array}{rl}
    \textrm{Var}[Y_i(x)]=&\E_{X_i}[W_i^2(x)]-\{\E_{X_i}[W_i(x)]\}^2\\
    \le&\int_{\Om} K_t^2(x,y)f^2(y)p(y)dy + [\int_{\Om}
    K_t(x,y)f(y)p(y)dy]^2\\
    \le & \frac{1}{t^{d/2}}M^2 b \int_{\Om} K_t(x,y)dy + M^2b^2C_g^2\\
    \le & \frac{1}{t^{d/2}}M^2 b C_g
    + M^2b^2C_g^2=(C_v+M^2b^2C_g^2t^{d/2})/t^{d/2}\\
\end{array}
\end{equation}
where $C_v=M^2 b C_g$. By the Bernstein's inequality
\begin{equation}
    P\left(\left|\frac{1}{n}\sum_{i=1}^n Y_i(x)\right|>\epsilon \right)\le 2
\exp{\left(-\frac{nt^{d/2}\epsilon^2/2}
    {(C_v+M^2b^2C_g^2t^{d/2})+(C_m+Mb C_g t^{d/2})\epsilon/3}\right)}
\end{equation}
\end{proof}

\emph{Proof of theorem \ref{thm:complexity}}: For $x\in \overline{\Omega}$, we put $f(x)-f(X_i)$ into Lemma~\ref{lemma:convergencerate}, obtaining:
\[\begin{split}
&P\left(\left|L_{n,t}f(x)-L_tf(x)\right|>\epsilon\right)\\
=&P\left(\left|\frac{1}{n}\sum_{i=1}^n
    K_t(x,X_i)(f(x)-f(X_i))-\E_Z[K_t(x,Z)(f(x)-f(Z))]\right|>t\epsilon \right)\\
\le &2\exp{\left(-\frac{nt^{(d+4)/2}\epsilon^2}
    {2C_v+2C_mt\epsilon/3}\right)}
    \end{split}
\]
The proof of theorem \ref{thm:complexity} is done.

\section{Experimental results}
\label{appendix:sec:exp}

Below we provide some numerical results validating  theoretical analysis, some experiments with real datasets and several numerical experiments in support of our conjectures on the eigenfunction behavior.

\subsection{A numerical example}
\begin{figure}[ht]
\vskip -0.2in
\begin{center}
\subfigure[Sample data from $\Om$]{
\includegraphics[width=0.32\columnwidth]{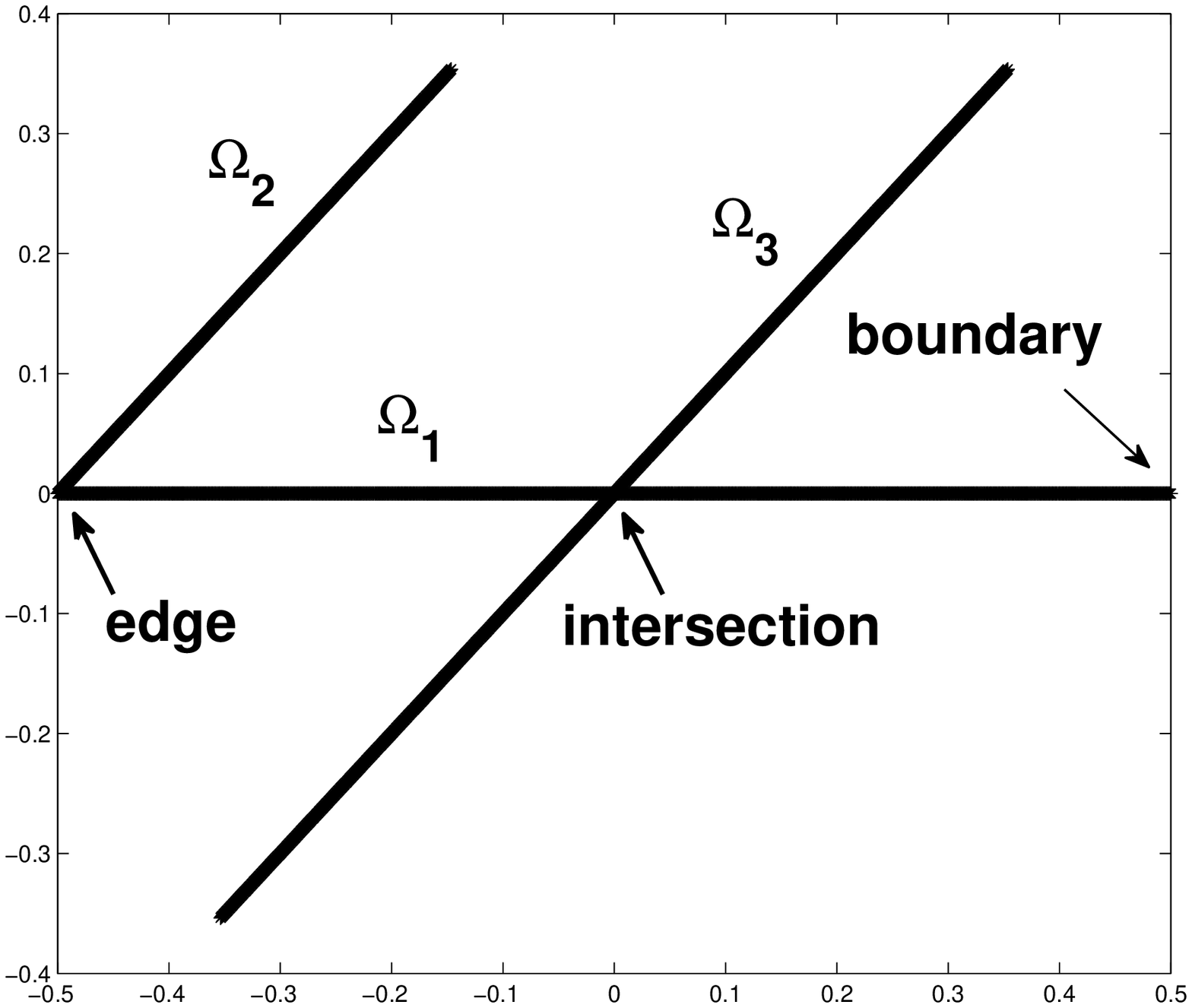}
}\subfigure[$\myL_{n,t}f(x)$ over $\Om_1$]{
\includegraphics[width=0.32\columnwidth]{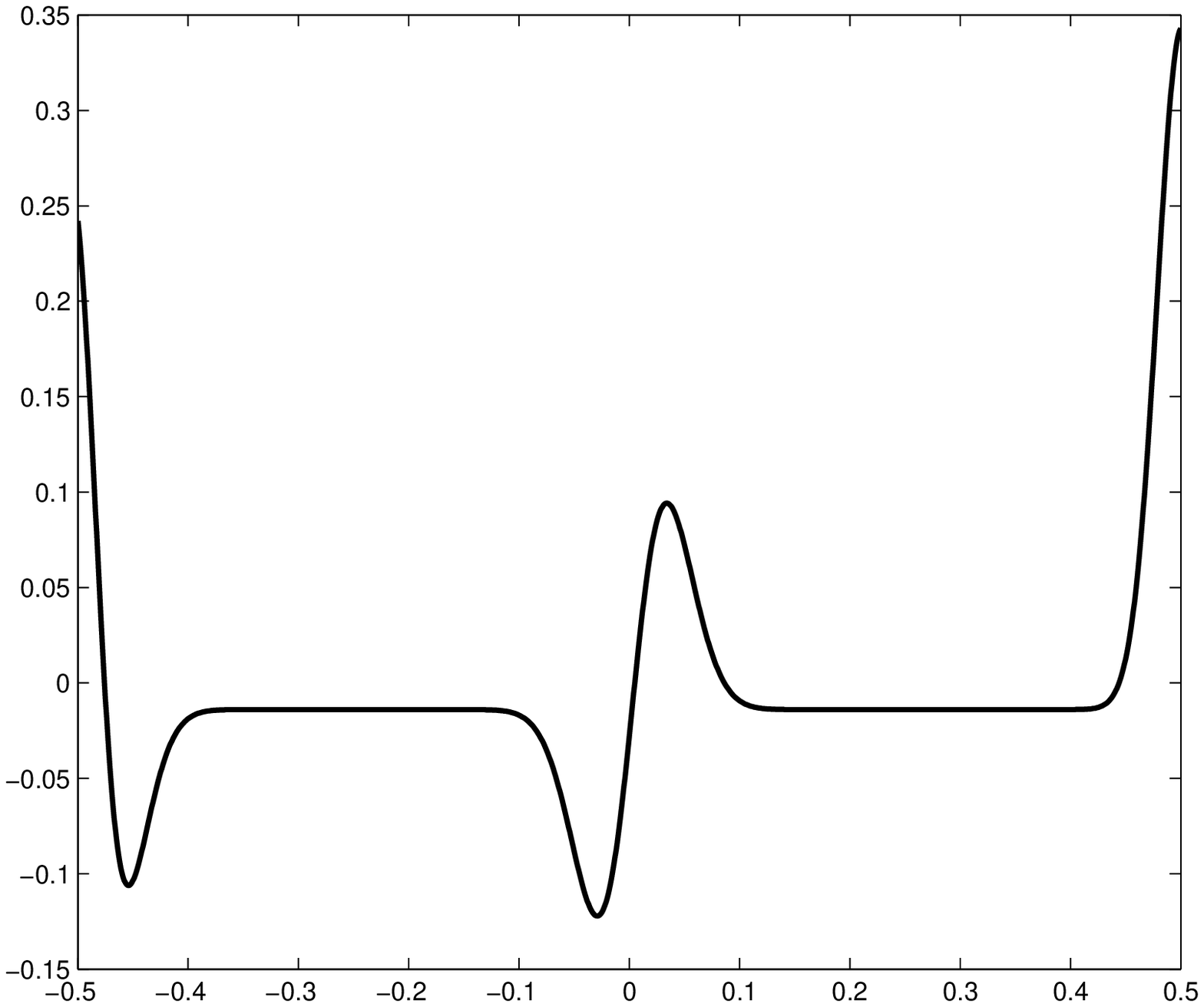}}
\subfigure[
$\log(|\myL_{n,t}f(x)|)$ vs $\log(t)$]{
\includegraphics[width=0.32\columnwidth]{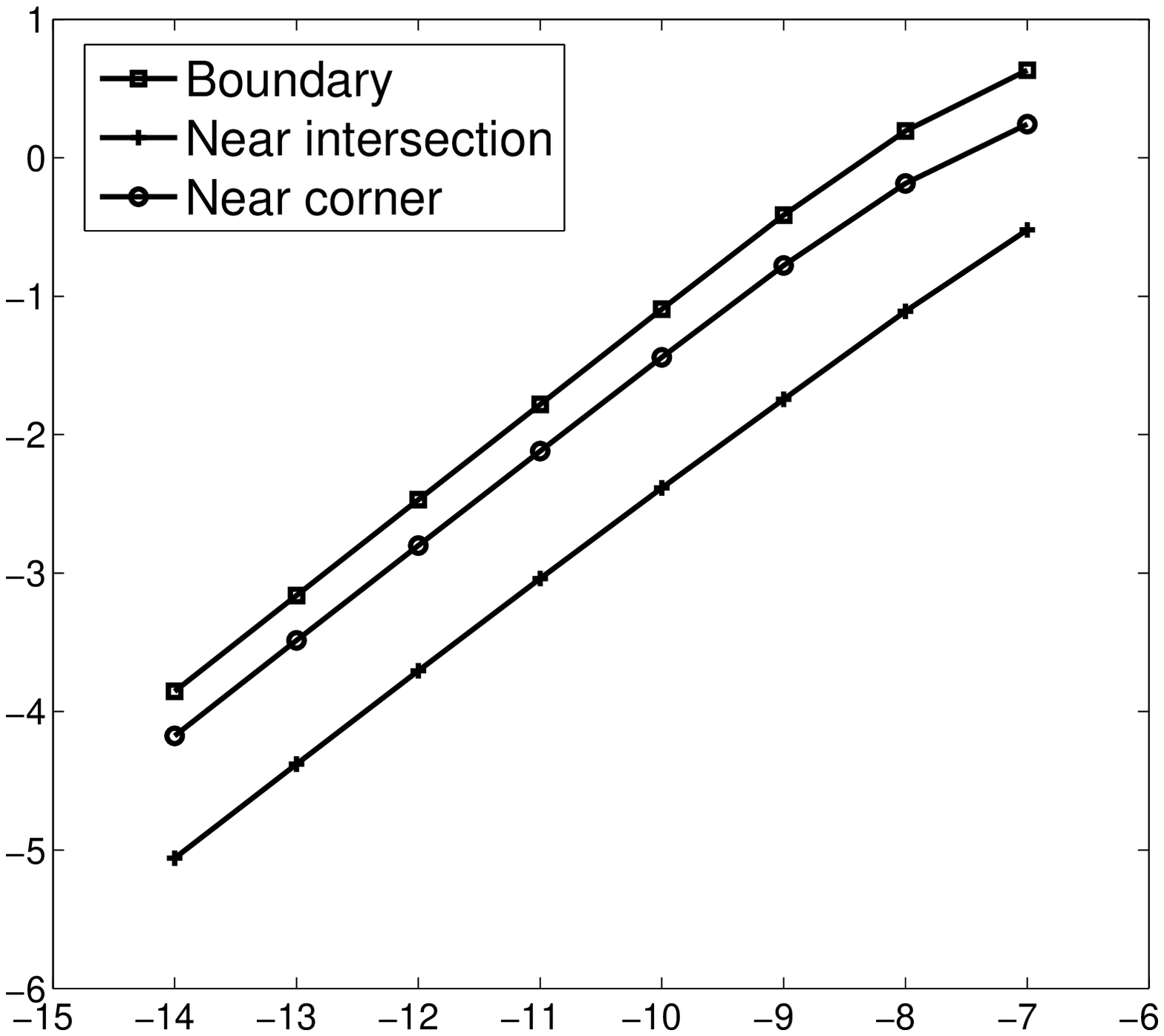}}
\vskip -0.1in\caption{$\myL_{n,t}f(x)$ with
$f(x_1,x_2)=(x_1+0.2)^2+x_2^2$ over $\Om$. In (c), $x$ axis is
$\log(|L_{n,t} f(x)|)$ and $y$ axis is $\log t$. The three curves
are for three points near each type of singularity.} \vskip -0.1in
\label{fig:ScaleBehavior}
\end{center}
\end{figure}

In this section, we illustrate the behavior of graph Laplacian on or near the singularity points through a simple numerical example. In order to explore different cases, we consider the union of 1-dimensional manifolds $\Om$ as shown in Figure~\ref{fig:ScaleBehavior} (a), which is a combination of three linear intervals, $\Om_1$, $\Om_2$, and $\Om_3$ in $\R^2$. We take 2500 point uniformly spaced over the singular manifold and choose $f(x_1,x_2)=(x_1+0.2)^2+x_2^2$ (restricted to the manifold).

{\bf Shape and Scaling Behavior:} The value of $\myL_{n,t}f(x)$ over all points in $\Om_1$ is shown in Figure~\ref{fig:ScaleBehavior} (b). We can see that $\myL_{n,t}f(x)$ near the boundary is approximately half a Gaussian function, while the function near the intersection is similar to a function of the form $\varphi(x) = a x e^{-bx^2}$, as predicted by Theorem~\ref{thm:intersection}. The shape of $\myL_{n,t}f(x)$ near the corner is also what we expect from Theorem~\ref{thm:edge}.

To explore the scaling effects, in Figure \ref{fig:ScaleBehavior} (c) we plot values of $\log{|\myL_{n,t}f(x)|}$ against $\log(t)$ for different values of $t$. Each graph corresponds to a fixed point $x$ chosen near a boundary, intersection or edge singularity. Our theoretical results predict a linear curve with slope $-\frac{1}{2}$ in $\log$-$\log$ coordinates (corresponding to $1/\sqrt{t}$, which is consistent with our experimental results as shown in (c).

\subsection{ Singularity Detection and Estimation}
\label{appendix:sec:expdetection}

\parpic[r]{\begin{tabular}{cc}
\includegraphics[height=3cm]{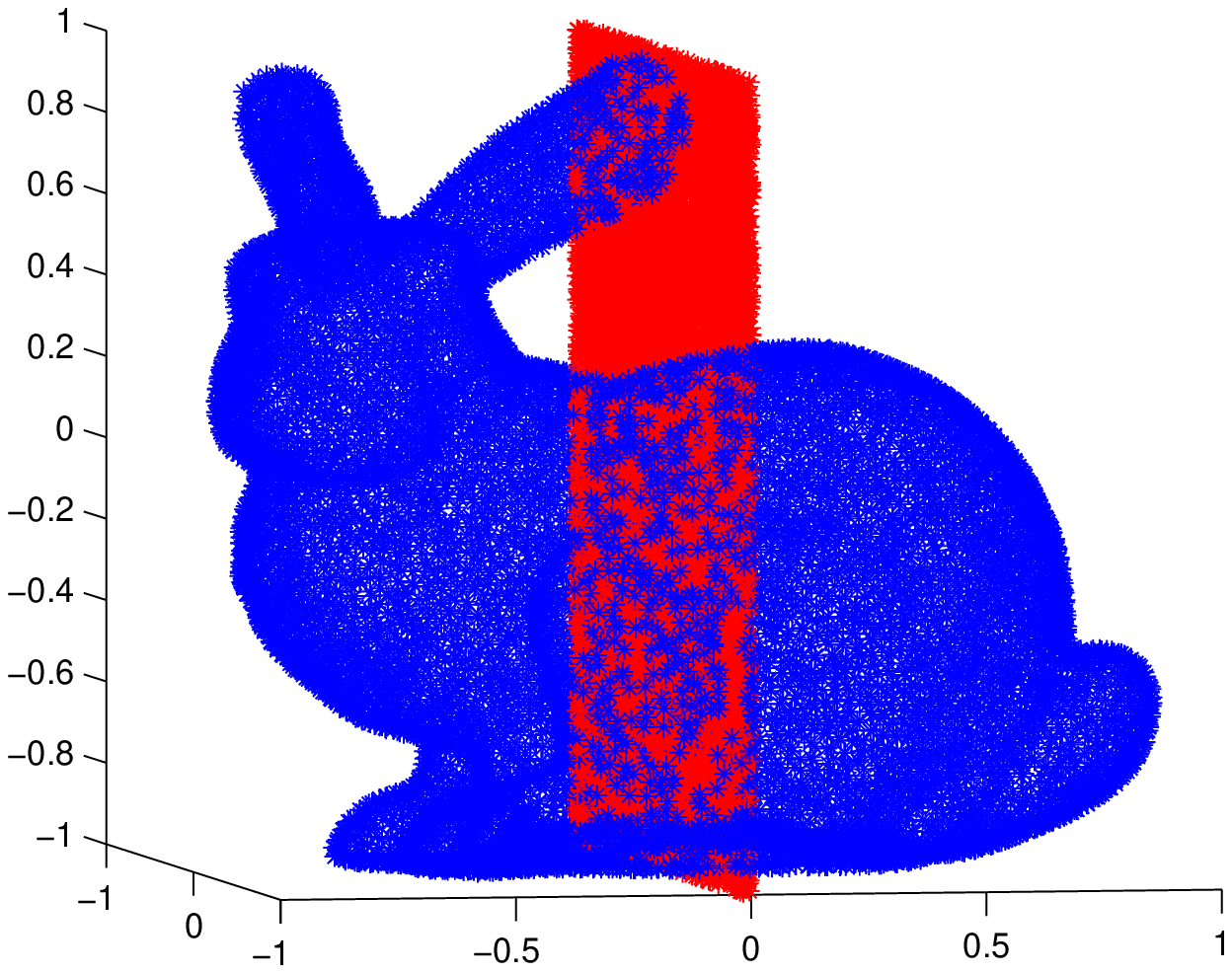} &
\includegraphics[height=3cm]{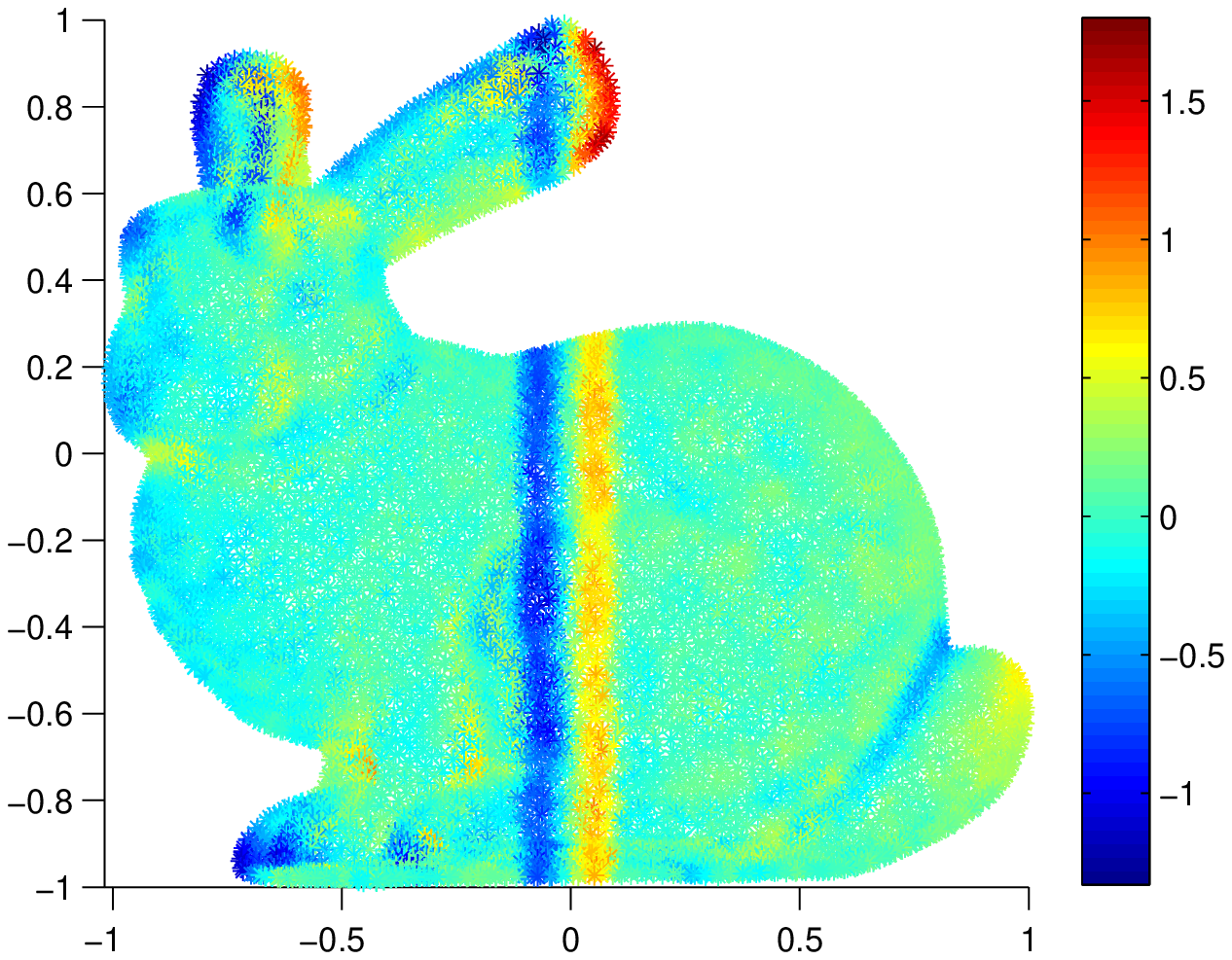} \\
(a) Data set & (b) Values of $\myL_{n,t} f$\\
  & on Bunny when $f(x,y,z)=y$.\\
\end{tabular}
}
\paragraph{Bunny Model:} In the following example, we use the union of the Stanford bunny dataset (in blue) and a plane $y=0$ (in red), which intersects the bunny in the middle as shown in panel (a). We have take $10000$ points for both the bunny and the plane in the point cloud, and choose the function $f=y$. The values of $\myL_{n,t} f$ on the bunny are shown in panel (b). We can see that the values of $\myL_{n,t} f$ at the intersection are $0$ and values near the intersection go up (in yellow) and down (in blue) as expected from the theoretical analysis of the intersection singularity.

\begin{figure}
\begin{center}
\subfigure[]{
\includegraphics[width=0.40\columnwidth]{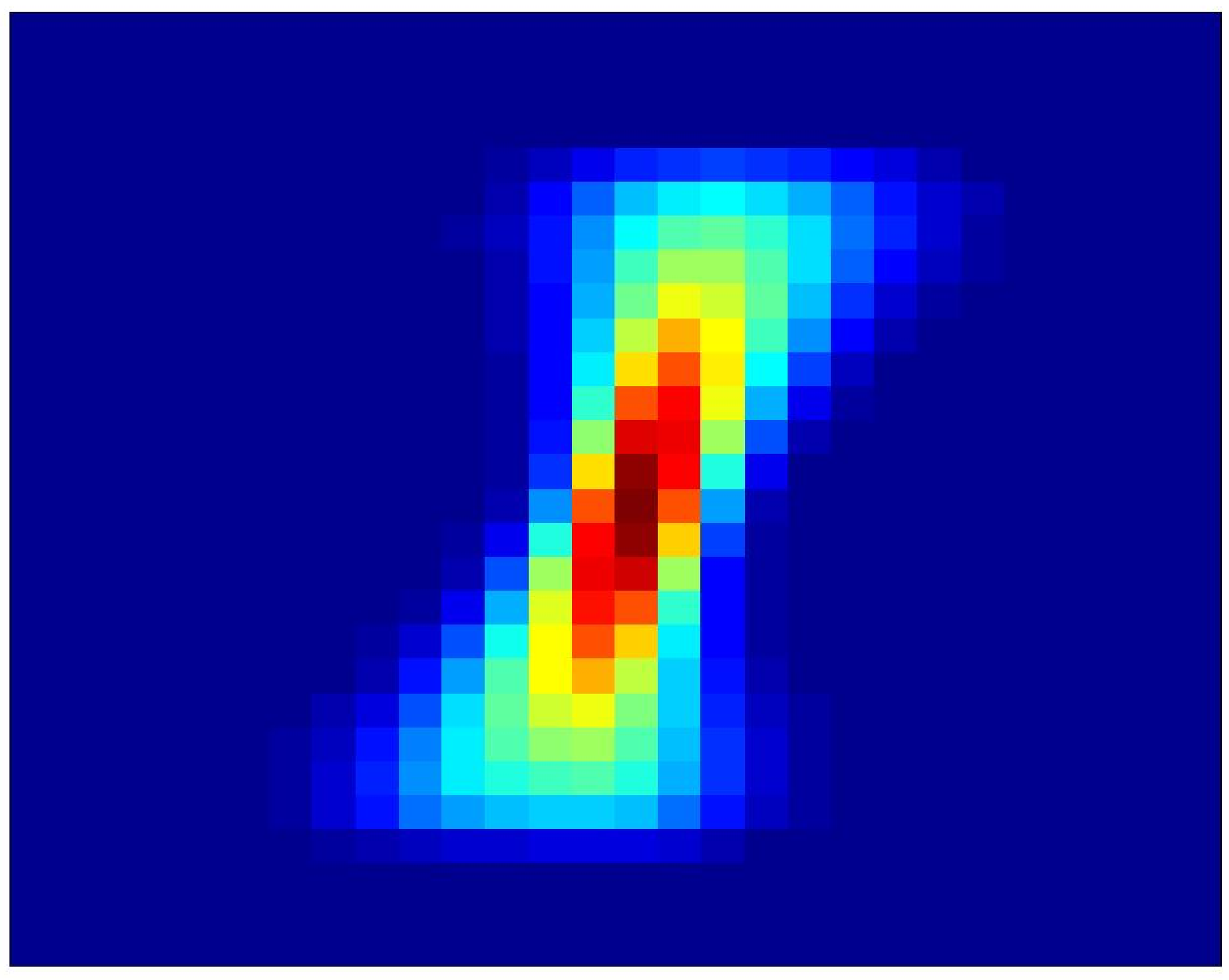}}
\subfigure[]{
\includegraphics[width=0.40\columnwidth]{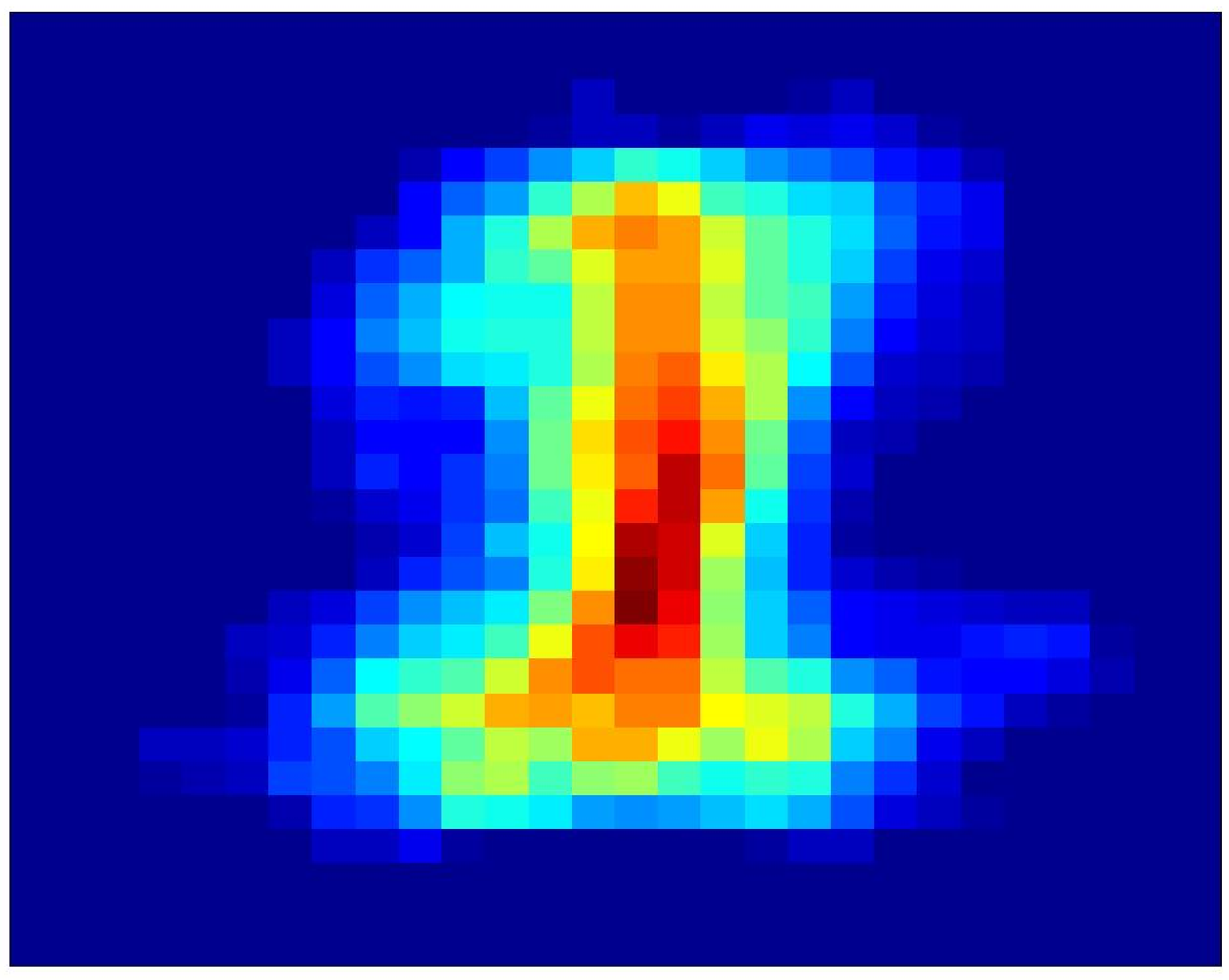}}
%
%
%
%


\caption{An averaged image of digits "1" vs an averaged image with a large value of graph Laplacian constructed from the data set of  images,  applied to a function defined by the sum of pixel intensities.}\label{fig:bd:MNIST}
\end{center}
\end{figure}

\paragraph{MNIST Data:} From the scaling behavior of the graph Laplacian near singularities, it is potentially possible to detect certain such singularities on real world data sets.
We use the example of the MNIST  digit images. The image of each digit can be thought of as a sample from a low-dimensional manifold embedded in the high-dimensional pixel space. We choose function $f(x)$ to be the summation of the pixel intensity of each image. We take the images corresponding to the top $2\%$ of the highest value of $\myL{n,t} f(x)$ as the ``near singularity/boundary'' images, which are compared to the ``average'' images of each digit.
The results are shown in Figure~(\ref{fig:bd:MNIST}). We see that the images on the right are significantly larger (and systematically different) from average, suggesting that they may belong to the part of the boundary, which can be detected through the sum-of-pixel-intensity function.

%

\subsection{Laplacian Eigenfunctions on  Singular Manifolds}
\label{appendix:sec:eigenexp}

In this section, we will provide two numerical examples to validate our to conjectures about the Laplacian eigenfunctions on the singular manifolds.

\begin{figure}
\begin{center}
\subfigure[$\om_1$]{
\includegraphics[width=0.40\columnwidth]{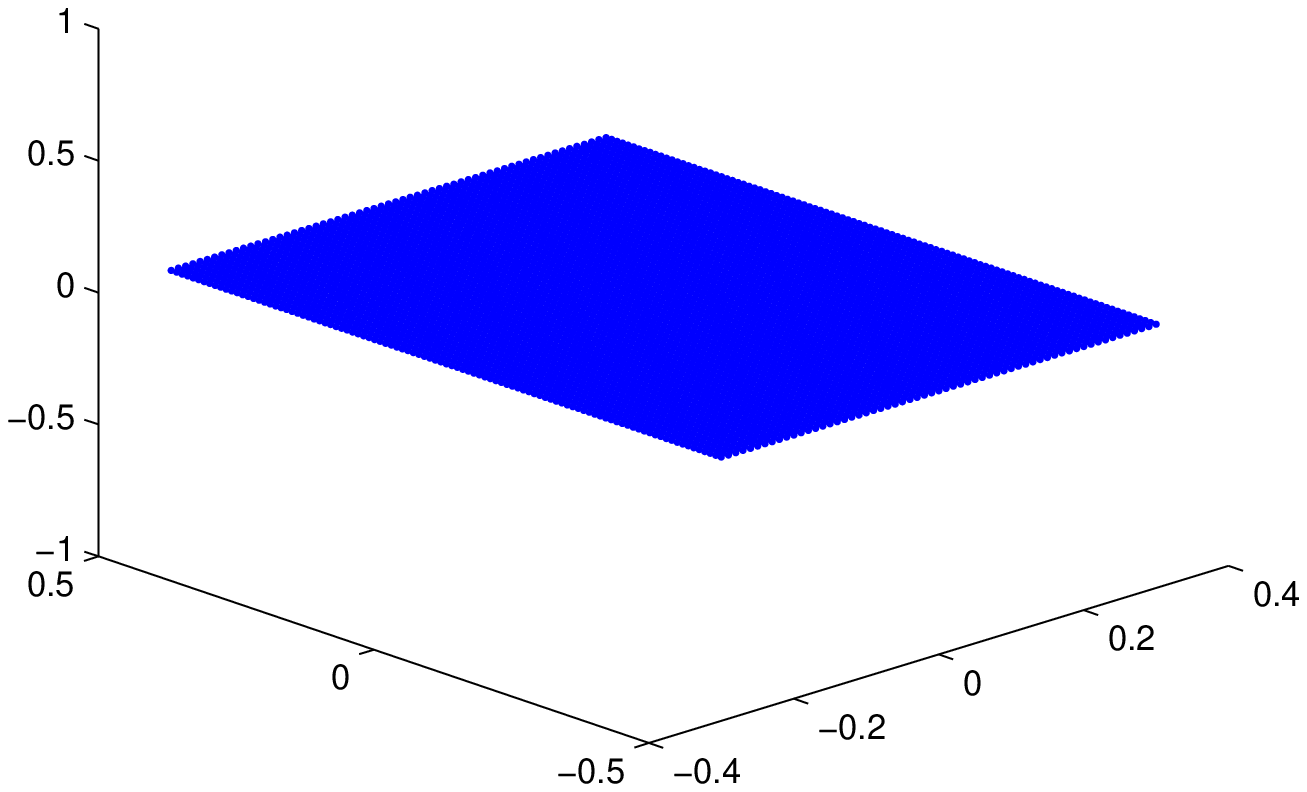}}
\subfigure[$\om_2$]{
\includegraphics[width=0.40\columnwidth]{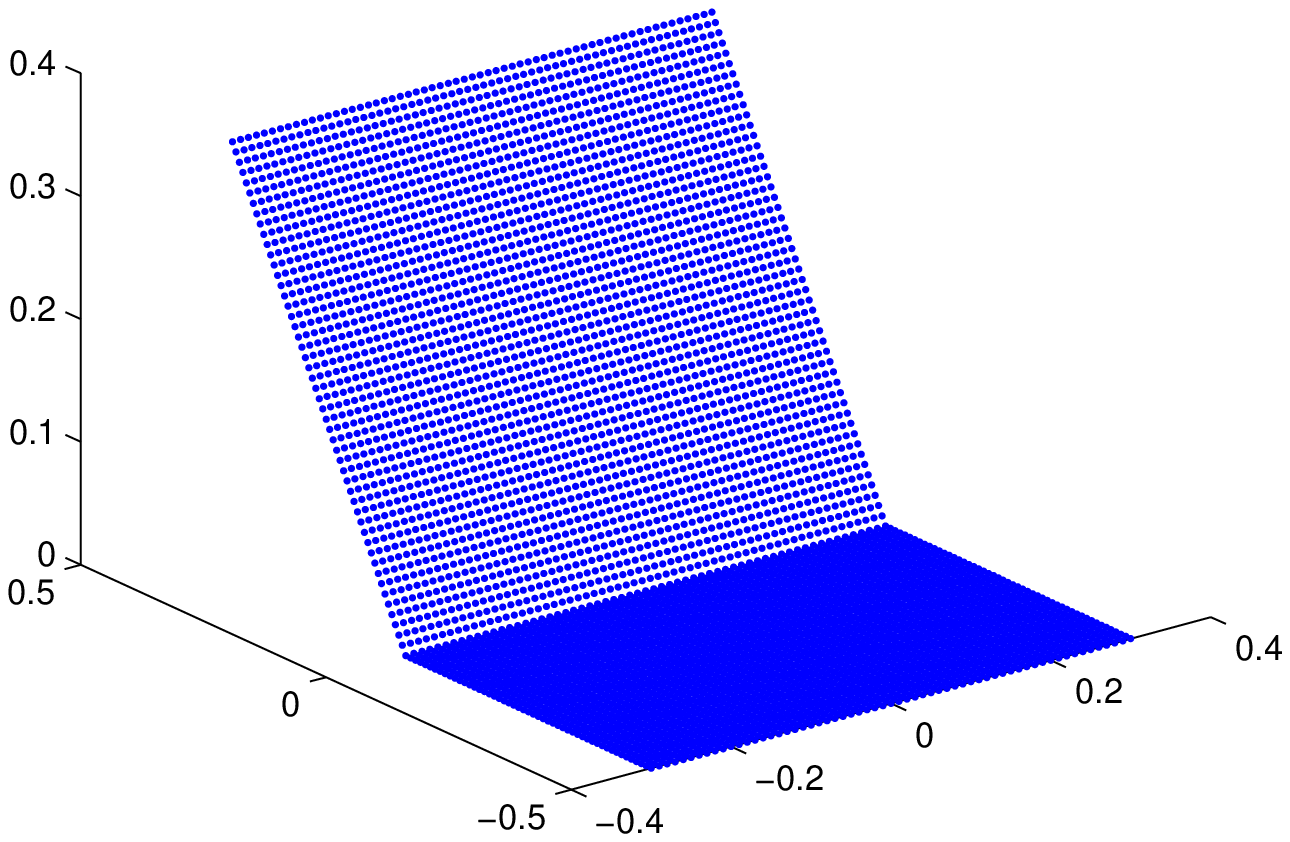}}
\caption{Rectangle and folded rectangle $\om_1$ and $\om_2$.}\label{fig:data_rect}
\end{center}
\end{figure}

\paragraph{Folded Rectangle:} In this example, we consider two manifolds: (1) the rectangular region $\om_1$, $(-0.3,0.3)\times (-0.5,0.5)\times \{0\}$, in $\R^3$ and  (2) the "folded" rectangle $\om_2$, obtained by transforming  the positive $y$ part  of $\om_1$, i.e. $\{(x,y,z)\in\om_1: y>0\}$, by applying the linear transformation given by the matrix
\[
\begin{bmatrix}
1 & 0 & 0\\
0 & \cos(\pi/4) & -\sin(\pi/4) \\
0 & \sin(\pi/4) & \cos(\pi/4)
\end{bmatrix}
\]
and keeping  the rest of $\om_1$ fixed as showing in Fig.~\ref{fig:data_rect}.
We see $\om_1$ and $\om_2$ are intrinsically isometric, yet $\om_2$ has an edge singularity.

We choose $6000$  uniformly-spaced points in  both  $\om_1$ and $\om_2$.

\begin{figure}
\begin{center}
\subfigure[$\om_1$]{
\includegraphics[width=0.40\columnwidth]{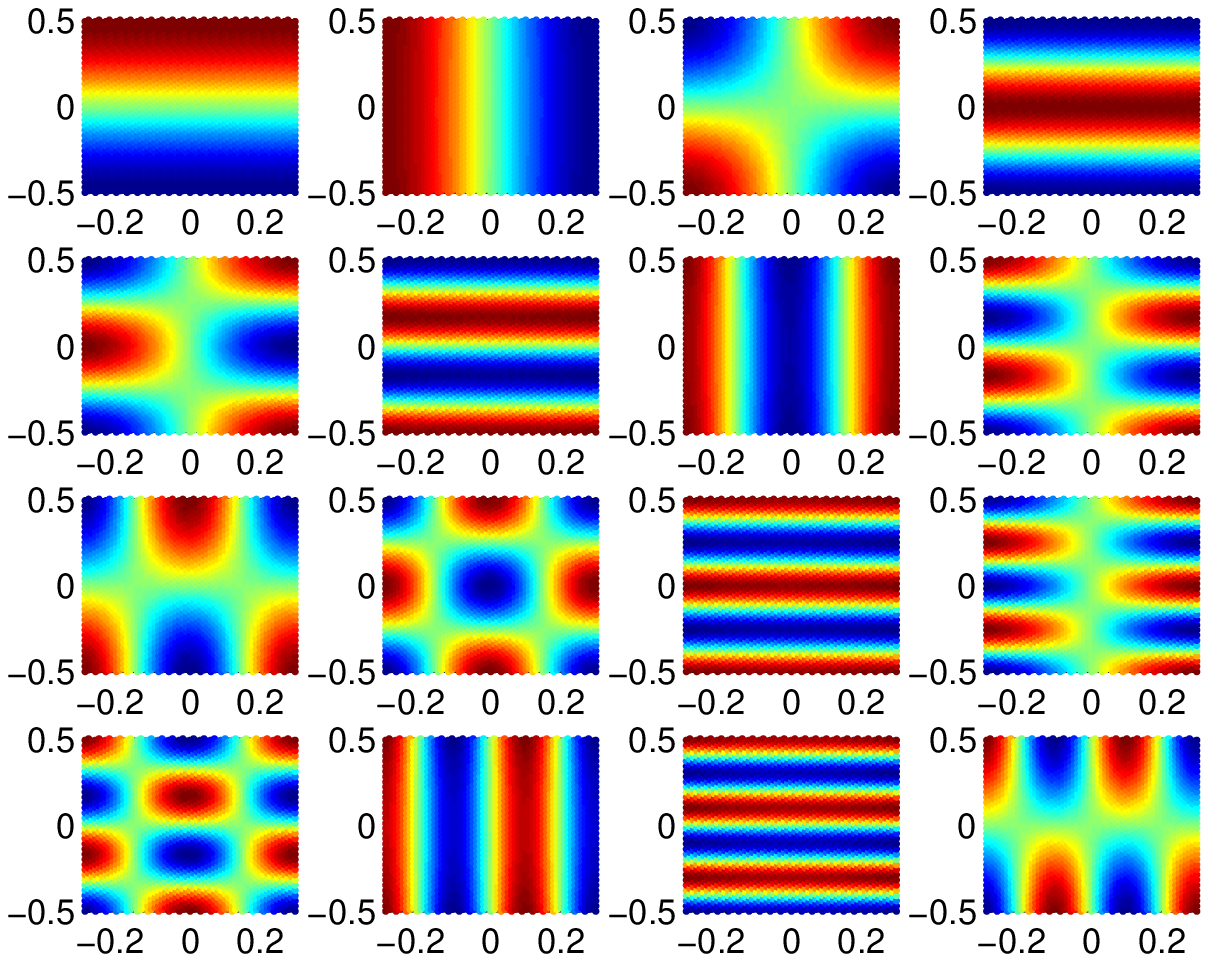}}
\subfigure[$\om_2$]{
\includegraphics[width=0.40\columnwidth]{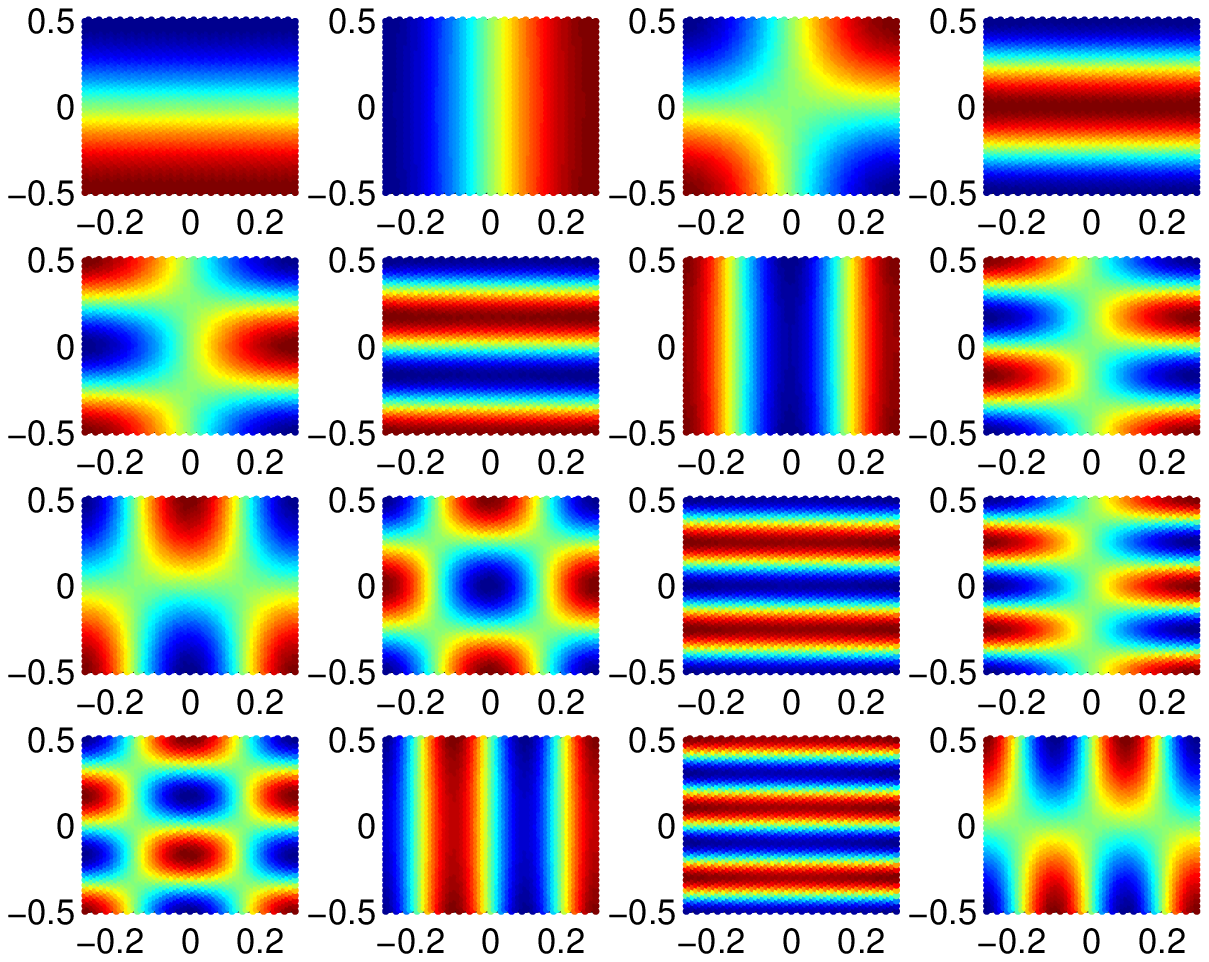}}
\caption{2nd-17th eigenvectors of the graph Laplace matrices of the two manifolds $\om_1$ and $\om_2$.}\label{fig:eigen_rect}
\end{center}
\end{figure}

With this two data sets, we construct  their graph Laplace matrices respectively, and calculate their eigenvalues and eigenvectors using the Gaussian kernel with $t=10^{-4}$. Figure \ref{fig:eigen_rect} shows the first 16 nontrivial eigenvectors of two graph Laplace matrices. We can see that the corresponding eigenvectors match nearly perfectly.
It can be easily verify that they also match eigenfunctions of the rectangle with Neumann boundary conditions.

To give some precise numerical results we measure the relative difference  between  the sets of eigenvalues.  Let $\lambda_k^1$ and $\lambda_k^2$ be the two vectors of first $k$ eigenvalues for $\om_1$ and $\om_2$ respectively. We define the  (relative) difference as
\[
\text{diff}_k = \frac{\|\lambda_k^1-{\lambda}_k^2\|}{\|\lambda_k^1\|}
\]
The table below shows the difference for various values of $k$. We observe that the approximation appears quite precise with  difference  of the order of $0.1\%$, thus providing further evidence in support of our conjecture on the behavior of eigenfunctions.
\begin{center}
\begin{tabular}{|c|c|c|c|}
\hline
k & 10 & 50 & 100 \\ \hline
$\text{diff}_k$ & 0.0014 & 0.0012 & 0.0011 \\  \hline
\end{tabular}
\end{center}


\end{document}